\documentclass[11pt]{article}
\usepackage[final]{acl}
\usepackage{amsmath,amsfonts,bm}

\def\eqref#1{equation~\ref{#1}}
\def\1{\bm{1}}

\DeclareMathAlphabet{\mathsfit}{\encodingdefault}{\sfdefault}{m}{sl}
\SetMathAlphabet{\mathsfit}{bold}{\encodingdefault}{\sfdefault}{bx}{n}

\usepackage{graphicx}
\usepackage{subcaption}
\usepackage{hyperref}
\usepackage{url}

\usepackage{soul}
\usepackage{xcolor}   % For coloring
\usepackage{tcolorbox} % For boxed background

\definecolor{peachpink}{RGB}{241, 208, 202}
\definecolor{lightyellow}{RGB}{244,244,152} % Experts (light-yellow boxes)
\definecolor{lightgreen}{RGB}{176,228,179} % MoICL (light-green box)
\definecolor{palecyan}{RGB}{190,228,243} % Partitions (light-blue boxes)
\definecolor{cyan}{RGB}{214, 223, 252} % Demonstrations D (cyan box)
\definecolor{lightgreenX}{RGB}{194, 247, 201}

\newcommand{\hlpeach}[1]{{\sethlcolor{peachpink}\hl{#1}}}
\newcommand{\hlyellow}[1]{{\sethlcolor{lightyellow}\hl{#1}}}

\newcommand{\hlgreenX}[1]{{\sethlcolor{lightgreenX}\hl{#1}}}
\newcommand{\hlcyan}[1]{{\sethlcolor{palecyan}\hl{#1}}}
\newcommand{\hlaqua}[1]{{\sethlcolor{cyan}\hl{#1}}}

\usepackage{times}
\usepackage{latexsym}

\usepackage[T1]{fontenc}
\usepackage[utf8]{inputenc}

\usepackage{microtype}

\usepackage{inconsolata}

\usepackage{caption}
\usepackage{wrapfig}
\usepackage{subcaption}

\usepackage{graphicx}
\usepackage{booktabs}
\usepackage{adjustbox}
\usepackage{multirow}
\usepackage{amsmath}
\usepackage{paralist}
\usepackage{enumitem}

\usepackage[T1]{fontenc}
\usepackage[utf8]{inputenc}

\usepackage{hyperref}
\usepackage{xspace}

\newcommand*{\eg}{e.g.\@\xspace}
\newcommand*{\ie}{i.e.\@\xspace}

\usepackage[capitalise]{cleveref}

\usepackage[normalem]{ulem}
\useunder{\uline}{\ul}{}

\definecolor{darkblue}{rgb}{0.0, 0.0, 0.5}

\hypersetup{
    colorlinks=true,
    linkcolor=darkblue,      % Color for internal links (sections, pages)
    citecolor=darkblue,      % Color for citation links (bibliography)
    urlcolor=darkblue        % Color for external links (URLs)
}

\newcommand{\moicl}{\textsc{MoICL}\xspace}
\newcommand{\hypernet}{\emph{adaptive}\xspace}
\newcommand{\partstatic}{\textbf{Static}\xspace}
\newcommand{\partrandom}{\textbf{Random Size}\xspace}

\definecolor{lightred}{RGB}{255, 204, 204}

\def\std#1{\scriptsize$\pm$#1}

\author{
\normalsize
Giwon Hong$^1$\quad
Emile van Krieken$^1$\quad
Edoardo M. Ponti$^1$\\
\normalsize
\textbf{Esmeralda S. Whitammer}$^1$\quad
\textbf{Pasquale Minervini}$^{1,2}$
\\ 
\normalsize
$^1$University of Edinburgh, United Kingdom \qquad
$^2$Miniml.AI, United Kingdom\\
\normalsize
\texttt{\{giwon.hong, p.minervini\}@ed.ac.uk}
}

\title{Mixtures of In-Context Learners
}

\begin{document}
\maketitle
\begin{abstract}
%
% LIMITATIONS
In-context learning (ICL) adapts LLMs by providing demonstrations without fine-tuning the model parameters; however, it is very sensitive to the choice of in-context demonstrations, and processing many demonstrations can be computationally demanding.
%
% it does not differentiate between demonstrations and quadratically increases the complexity of Transformer LLMs, exhausting the memory.\pmm{?}
% CONTRIBUTION
%As a solution,
%
We propose Mixtures of In-Context Learners (\moicl), a novel approach that uses subsets of demonstrations to train a set of experts via ICL and learns a weighting function to merge their output distributions via gradient-based optimisation.
%based on a training set. %\pmm{subsets of demonstrations are not experts tho -- they are used to train experts}
%
In our experiments, we show performance improvements on 5 out of 7 classification datasets compared to a set of strong baselines (\eg, up to +13\% compared to ICL and LENS).
Moreover, we improve the Pareto frontier of ICL by reducing the inference time needed to achieve the same performance with fewer demonstrations.
Finally, \moicl is more robust to out-of-domain (up to +11\%), imbalanced (up to +49\%) and perturbed demonstrations (up to +38\%).\footnote{Our code is available at \url{https://github.com/HongGiwon/Mixture-of-In-context-Learners}} % or can filter these out from datasets.
% CONCLUSION
%
% Overall, \moicl is a more expressive approach to learning from demonstrations without exhausting the context window or memory.
%
\end{abstract}

\section{Introduction}
%
%In-context learning (ICL), where we condition a large language model (LLM) on a set of input-output examples (demonstrations) to perform a wide range of tasks~\citep{NEURIPS2020_1457c0d6, wei2022emergent}, is a transformative technique in NLP.
%
In-context learning (ICL) refers to the ability of a large language model (LLM) to learn to perform tasks by interpreting and adapting to examples (demonstrations) provided directly in the input context without requiring updates to its parameters~\citep{NEURIPS2020_1457c0d6, wei2022emergent}.
However, in ICL, the number of demonstrations may be limited by the memory requirements of the model~\citep{wei2022emergent}, and its effectiveness can vary significantly depending on which demonstrations are selected~\citep{lu-etal-2022-fantastically, chen2023relation} and how they are verbalised~\citep{DBLP:conf/acl/VoronovWR24}.
Current methods for selecting demonstrations are largely heuristic and do not adequately quantify the influence of individual examples on the generalisation properties of the model~\citep{DBLP:conf/naacl/LuWT0S24}.
\begin{figure}[t]
    \centering
    \includegraphics[width=\columnwidth]{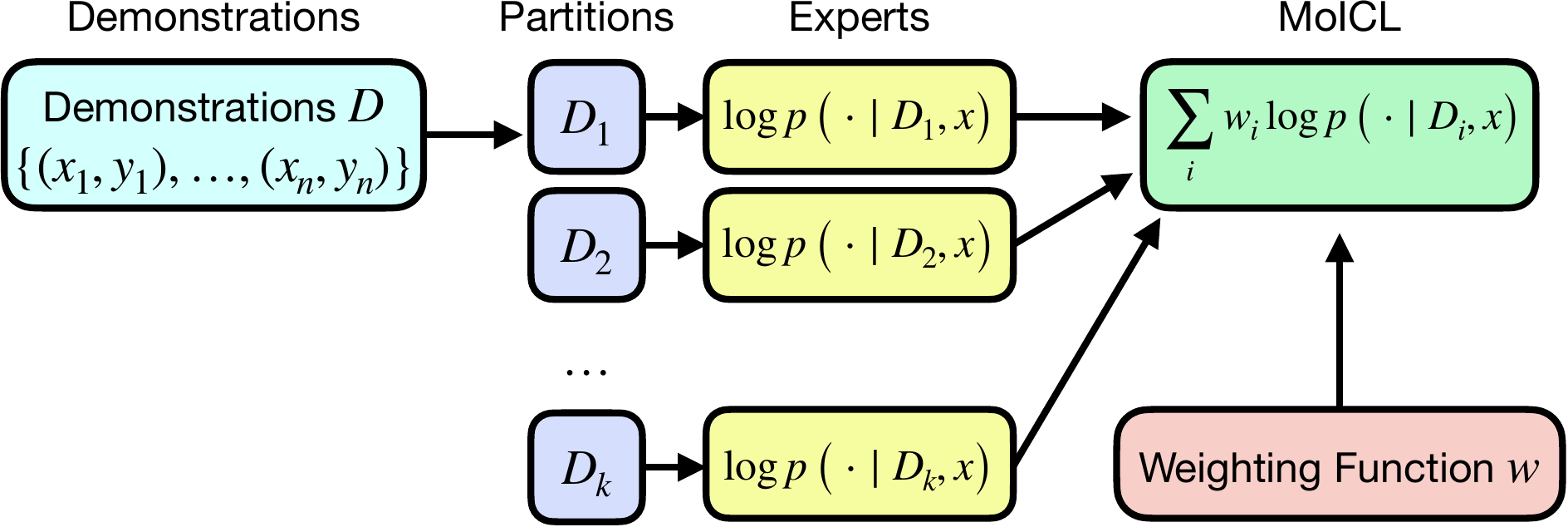}
    \caption{A \hlgreenX{Mixture of In-Context Learners} (\moicl) first partitions a \hlcyan{set of demonstrations} $D$ in $k$ \hlaqua{partitions} to create $k$ \hlyellow{experts trained via in-context learning}, and then combines their next-token predictions via a trainable \hlpeach{weighting function}. 
    }
    \label{fig:splash}
\end{figure}
Demonstrations are often selected randomly or based on simple heuristics~\citep{xu2024context}, which can lead to suboptimal results.

To address these issues, we propose Mixtures of In-Context Learners (\moicl), a method for dynamically learning how different sets of examples contribute to the prediction task.
\moicl prompts an LLM with multiple subsets of examples and combines their next-token distributions via a weighting function that can be trained via gradient-based optimisation methods; \cref{fig:splash} shows a high-level outline of the method.

We analyse the generalisation properties of \moicl in the following settings:
\begin{inparaenum}[(1)]
\item presence of out-of-domain (OOD) demonstrations, where some in-context demonstrations are sourced from a different dataset;
\item label imbalance, where the training label distribution is significantly skewed towards a subset of labels; and
\item perturbed demonstrations, where the labels of some demonstrations are perturbed to be completely incorrect.
\end{inparaenum}
\noindent In all cases, we find that \moicl produces significantly more accurate results than ICL.

Furthermore, \moicl does not require access to the internal parameters of the LLM, making it applicable to black-box LLMs, and it significantly reduces the computational complexity issues arising from long contexts since it allows the distribution of the training samples among multiple experts.
We also show that \moicl can be made more efficient by sparsifying the mixing weights.

We summarise our contributions as follows: 
\begin{itemize}[left=0pt,nosep]
\item We introduce Mixtures of In-Context Learners (\moicl), %which assigns weights to each demonstration subset and learns from them, 
a method for combining the predictions of multiple models trained via ICL and identifying the optimal mixture weights via gradient-based optimisation. 
%(\cref{sec:MOICL}) % \pmm{ref to corresponding sections for each bullet point?}
(\S \ref{sec:MOICL})
\item We show that \moicl is resilient to perturbed demonstrations and label imbalance. 
%(\cref{sec:experiments})
(\S \ref{sec:experiments})
\item We demonstrate that \moicl is competitive with standard ICL while being significantly more \hbox{data-,} memory-, and compute-efficient. (\S \ref{ssec:data_eff})
\end{itemize}
%

\iffalse
\begin{figure*}[t]
    \centering
    \includegraphics[width=\linewidth]{figures/figure2_ver3.pdf}
    \caption{Illustration of our proposed Mixture of In-Context Learners (\moicl), compared to the standard In-context Learning. In \moicl, we first partition $n$ demonstrations into $k$ subsets, and each subset is used to train an \emph{expert} via ICL. Then, the next-token distributions of the resulting experts are combined by a trainable \emph{weighting function} $w$ to produce a single next-token distribution for the \moicl.}
    \label{fig:main_figure}
\end{figure*}
\fi

\section{Mixtures of In-Context Learners}
\label{sec:MOICL}
\subsection{In-Context Learning}
\label{ssec:ICL}

In ICL, given a LLM with next-token distribution $p(\cdot)$, a set of $n$ demonstrations $D$ = \{$(x_{1}, y_{1}), \ldots, (x_{n}, y_{n})$\} and an input instance $x$, the model generates a response $y$ when prompted with the concatenation of the examples in $D$ and $x$:
\begin{equation} \label{eq:posterior_from_concat}
\begin{aligned}
y & \sim p(y \mid  x_1, y_1, \dots, x_n, y_n, x) \\
& = p(y \mid D, x),
\end{aligned}
\end{equation}
\noindent we refer to the model in \cref{eq:posterior_from_concat} as \emph{concat-based ICL}~\citep{min-etal-2022-noisy}.
With concat-based ICL, given a demonstration set $D$, the model can generate a response $y$ for the input text $x$ without needing task-specific fine-tuning or access to the model parameters.
However, concat-based ICL is still problematic; recent works show that it is very sensitive to the choice of the prompts and in-context demonstrations~\citep{DBLP:conf/acl/VoronovWR24}, the number of demonstrations is bounded by the maximum context size~\citep{NEURIPS2020_1457c0d6}, and, in Transformer-based LLMs, the cost of self-attention operations grows quadratically with the number of in-context samples~\citep{liu2022few}.

\subsection{Mixtures of In-Context Learners} \label{sec:moicl}
We propose Mixtures of In-Context Learners (\moicl), a method for addressing the limitations of concat-based ICL (\S \ref{ssec:ICL}).
We first partition the set of demonstrations $D$ into $k$ disjoint subsets $D_{1}, \ldots, D_{k}$, \ie $D = D_{1} \sqcup \ldots \sqcup D_k$.\footnote{We compare different partitioning strategies in \cref{ssec:partition}}
%
%\begin{equation}
%\label{eq:part}
%
%D = D_{1} \sqcup D_{2} \sqcup \ldots \sqcup D_k.
%
%\end{equation}
%
Then, each demonstration subset $D_{i} \subseteq D$ is passed to the LLM along with the input text $x$, and we denote the LLM predictors with the different demonstrations in their context as \emph{experts}.
The next-token distributions of the experts are combined using a vector of mixing weights $w \in \mathbb{R}^{k}$:
\begin{equation} \label{eq:moicl}
\hspace{-2pt} p\left(y \mid D, x\right) \propto \exp\left[\sum_{i=1}^{k} w_i \log p\left(y\mid D_i,x\right)\right]
\end{equation}
\noindent where each $w_i \in \mathbb{R}$ represents the contribution of the expert denoted by $p\left(y\mid D_i,x\right)$ to the final next-token distribution $p\left(y \mid D, x\right)$, and each $p\left(y\mid D_i,x\right)$ is a concat-based ICL expert, as in \cref{eq:posterior_from_concat}, with its own set of demonstrations.~\footnote{The formulation in \cref{eq:moicl}  uses a product of experts; it is also possible to use a regular mixture of experts --- we experimentally compare them in \cref{fig:figure3} and \cref{appendix:mixing}.}
%

%
%\paragraph{Mixture Weighting Function}
%
For computing the weights $w \in \mathbb{R}^{k}$ of the $k$ experts in \moicl, we use a vector of trainable parameters $w \in \mathbb{R}^{k}$, where $w_i$ denotes the weight associated to the $i$-th expert.
We refer to this parameterisation of the weighting function as \emph{scalar weights} (\emph{scalar}).
%

%
%\subsection{Mixture Weighting Functions}
%
%We consider the following weighting functions for calculating the weight $w_i \in \mathbb{R}$ of the $i$-th expert in \moicl:
%
%\begin{description}[leftmargin=5pt,nosep]
%
%\item[Scalar weights] -- we use a vector of trainable parameters $w \in \mathbb{R}^{k}$, where $w_i$ denotes the weight associated to the $i$-th expert. % The weights $w$ are initialised as $\forall i: w_{i} = 1/k$. \pmm{omit initialisation?}
%
\paragraph{Adaptive Mixture Weights}
In addition to a vector of parameters $w \in \mathbb{R}^{k}$, for computing the mixture weights in \moicl, we also experiment with using a hyper-network~\citep{DBLP:conf/iclr/HaDL17} $h_{\phi}(\cdot)$ with parameters $\phi$ to generate the weights of each expert $w_{i}$, given all in-context demonstration subsets concatenated, \ie $w_{1},\ldots,w_{k} = h_{\phi}(D_{1},\ldots,D_{k})$.
%
%\pmm{mention advantages? e.g. generalisation to unseen $D_1, \ldots, D_k$}
%
We learn the parameters of the weighting function $w$ by maximising the conditional log-likelihood of a training set $D_{T}$.
One advantage of using a hyper-network $h_{\phi}$ for dynamically computing the weights $w$ is that the model can provide weights for sets of demonstrations $D_{i}$ not seen during training.
We refer to this using of a hyper-network to generate weights as \hypernet \emph{weights} (\hypernet)
\paragraph{Sparsifying the Mixture Weights}
%
%\EK{It's not clear that this method is not standard applied to the method (I think it's only in 4.8?)}
One limitation of \moicl is that, for each token, it requires invoking the base LLM $k$ times, one for each expert with a different set of in-context examples.
To solve this issue, we propose to \emph{sparsify} the weighting coefficients $w \in \mathbb{R}^{k}$ so that only $k^{\prime}<k$ of them have non-zero values.
To achieve this, we define the output of the weighting function as $w = w^{\prime} \odot \text{top-}k^{\prime}(m)$, where $w^{\prime} \in \mathbb{R}^{k}$ are scalar weights for the $k$ experts, $m \in \mathbb{R}^{k}$ is a set of \emph{masking coefficients}, $\text{top-}k^{\prime}: \mathbb{R}^{k} \mapsto \{ 0, 1 \}^{k}$ is a function that produces a mask that selects the highest $k^{\prime}$ elements of a $k$-dimensional input vector, and $\odot$ is the element-wise product.
To back-propagate through the rationale extraction process, we use Implicit Maximum Likelihood Estimation~\citep[IMLE;][]{niepert2021implicit,DBLP:conf/aaai/Minervini0N23}, a gradient estimation method for back-propagating through continuous-discrete functions like $\text{top-}k^{\prime}$ into neural architectures.
%
%More specifically, let $\widehat{m} = \text{top-}k^{\prime}(m) \in \{ 0, 1 \}^{k}$ denote the $\text{top-}k^{\prime}$ mask.
%
%In our experiments using IMLE, we estimate the gradient of the loss w.r.t. the masking coefficients $\nabla_{m} \mathcal{L}$ as $\nabla_{m} \mathcal{L} \approx \text{top-}k^{\prime}(m) - \text{top-}k^{\prime}(m + \lambda \nabla_{\widehat{m}} \mathcal{L})$, where $\lambda \in \mathbb{R}_{+}$ is a hyperparameter.
%

\begin{table*}[t]
%\vspace*{-1em}
\centering
\begin{adjustbox}{width=1\textwidth}
\begin{tabular}{@{}lccccccc}
\toprule
\textbf{Method} ($n$ = 30) $\downarrow$ \textbf{Dataset} $\rightarrow$                & \textbf{Offensive} & \textbf{Hate} & \textbf{SST2} & \textbf{RTE}  & \textbf{FEVER} & \textbf{PAWS} & \textbf{QNLI}\\
\midrule
Concat-based ICL                    & 76.44\std{2.48}       & 53.54\std{4.29}  & 95.46\std{0.14}  & 86.43\std{1.26} &   80.63\std{0.49}  &   78.12\std{0.77}  & 89.08\std{0.44}\\
Similarity-based Search & 76.58\std{0.12}       & 57.02\std{0.02}  & 94.27\std{0.16}  & 82.24\std{0.48} &   81.62\std{0.11}  &   77.33\std{0.10}  & 88.60\std{0.10}\\
Random Search & 77.88\std{1.14} & 58.09\std{1.93} & \textbf{95.76\std{0.18}} & 86.57\std{1.43} & 82.13\std{0.10} & 78.88\std{0.57} & 89.99\std{0.26}\\
Plurality Voting & 73.21\std{0.60} & 58.51\std{0.78} & 94.06\std{0.20} & 75.38\std{2.18} & 79.42\std{0.42} & 64.97\std{1.02} & 88.98\std{0.17}\\
Ensemble-based ICL \citep{min-etal-2022-noisy}       &      73.35\std{0.44} & 53.68\std{4.27}  & 95.48\std{0.12}  &  86.43\std{1.34}  &   80.63\std{0.46} &  65.27\std{0.48} & 88.57\std{0.21} \\
\citet{le-etal-2022-shot} & 73.14\std{0.47} & 58.63\std{0.62} & 94.38\std{0.19} & 76.32\std{1.89} & 79.40\std{0.43} & 65.29\std{0.50} & 88.54\std{0.23}\\
LARA \cite{huang2025divide} & 74.28\std{0.44} & 49.72\std{2.06} & 94.01\std{0.51} & 79.78\std{1.46} & 78.84\std{1.62} & 73.04\std{0.62} & 87.36\std{0.94}\\
LENS \cite{li2023finding} & 78.70\std{0.67} & 53.20\std{3.11} & 93.81\std{0.16} & 84.98\std{0.74} & 80.07\std{0.29}  & 75.60\std{0.72} & 89.04\std{0.40}\\
PEFT~\citep[LoRA,][]{hu2022lora} & \it 79.79\std{4.07} & 53.76\std{4.98} & 85.89\std{6.32} & \textbf{88.88\std{2.78}} & 59.78\std{0.62} & 54.82\std{3.08}&57.24\std{4.77}\\
\midrule
Mixture of ICL (\emph{uniform})   & & & & & & &\\
\quad $k=5$   & 73.77\std{1.60} & 59.29\std{1.23}  & 95.39\std{0.30}  & 83.10\std{1.28} &  80.12\std{0.64} &  75.37\std{0.53} & 89.65\std{0.22} \\
\quad $k=10$   & 74.00\std{0.87} & 61.70\std{1.61}  & 94.91\std{0.19}  & 79.93\std{0.81} &  77.47\std{0.89} &  73.49\std{0.46} & 89.65\std{0.14}\\
\quad $k=30$   & 73.37\std{0.34}       & 59.12\std{0.47}  & 94.17\std{0.21}  & 77.26\std{1.02} &  79.46\std{0.36} &  65.29\std{0.51} & 88.66\std{0.25}\\
%\midrule
Mixture of ICL (\emph{scalar})   & & & & & & & \\
\quad $k=5$   & 78.35\std{1.49} & \it 66.03\std{3.31}  & 95.46\std{0.35}  & 84.12\std{1.07} &  81.43\std{0.90} &  77.56\std{0.53} & 89.99\std{0.44}\\
\quad $k=10$   & 79.42\std{1.48} & \textbf{66.52\std{2.62}}  & 95.32\std{0.27}  & 83.32\std{1.60} &  \it 82.04\std{0.98} &  \it 79.42\std{0.79} & \textbf{90.44\std{0.27}}\\
\quad $k=30$ & \textbf{81.33\std{0.69}}       & 63.45\std{1.69}  & 94.79\std{0.34}  & 79.93\std{0.93} &       \textbf{82.66\std{0.38}}   &       \textbf{79.50\std{0.33}}      & 90.11\std{0.20} \\
\bottomrule
\end{tabular}
\end{adjustbox}
\caption{Comparison between baseline methods and the proposed Mixture of In-Context Learners across classification tasks using Llama-3-8B-Instruct. $n$ and $k$ refer to the number of demonstrations and the number of subsets.
Bold text signifies the highest accuracy for each task. For baseline methods, $k$ is set to 30 when applicable.
}
\label{table:various_dataset}
\end{table*}

\section{Experiments}
\label{sec:experiments}

\paragraph{Datasets}
To study how well \moicl performs on classification tasks, we use the TweetEval~\citep{barbieri-etal-2020-tweeteval} offensive/hate, SST2~\citep{socher-etal-2013-recursive}, RTE~\citep{bentivogli2009fifth}, FEVER~\citep{thorne2018fever}, PAWS~\citep{zhang-etal-2019-paws}, and QNLI~\citep{wang2018glue} datasets.
We report the performance on the development set for SST2, RTE, FEVER, and QNLI.
For generation tasks, we use Natural Questions~\citep[NQ;][]{kwiatkowski2019natural} with an open-book setting~\citep{lee-etal-2019-latent}, GSM8K~\citep{cobbe2021gsm8k} employing Chain-of-Thought (CoT) reasoning. For the preference fine-tuning (\cref{appendix:DPO}), we use HH-RLHF~\citep{bai2022training}.

\paragraph{Models} We primarily used Llama-3-8B and its instruction-tuned model, Llama-3-8B-Instruct \citep{llama3modelcard} as our base LLMs. 
We use Llama-3-8B-Instruct for classification tasks, GSM8K, and HH-RLHF, while Llama-3-8B is used for NQ.
%We use Llama-3-8B for NQ, while Llama-3-8B-Instruct is used for classification tasks, GSM8K, and HH-RLHF; we use greedy decoding for generating from \moicl.
%
In \S \ref{ssec:moicl_proxy} (across multiple models), we used Llama-3.2-1B, Llama-3.2-1B-Instruct, and Llama-3.2-3B-Instruct.
We use Llama-2-7b-chat, 13b-chat, and 70b-chat \citep{touvron2023llama} for analysing the influence of model scale (\cref{sec:model_size}).
For longer context models in \cref{appendix:longer_context}, we use Llama-3.1-8B and Llama-3.1-8B-Instruct \citep{dubey2024llama}.
For hyper-networks, we used the T5 models ~\citep[T5-efficient-tiny/mini, t5-small, t5-base;][]{2020t5}
\paragraph{Baselines}
We compare \moicl with the following baselines. \textbf{Concat-based ICL} refers to the standard ICL introduced in ~\S \ref{ssec:ICL} where all demonstrations are concatenated into a single sequence and passed as input to the LLM along with the input text.
\textbf{Similarity-based Search} selects demonstrations from the demonstration pool according to their BM-25 similarity to the input text, concatenates them, and applies them in the same manner as Concat-based ICL.
\textbf{Random Search} samples random subsets from the demonstration pool, concatenates them and utilizes them as in Concat-based ICL.
Specifically, we sample $k$ random subsets and select the one that performs best on the training set. Here, $k$ is the maximum number of subsets used in \moicl, and the size of each subset is a random number between 1 and the number of demonstrations $n$. After finding the best subset, we evaluate it on the test set.
Inspired by ~\citet{wang2022rationale}, we also implement \textbf{Plurality Voting} baseline, which aggregates outputs through plurality voting over multiple inferences with different demonstrations.
\textbf{Ensemble-based ICL}~\citep{min-etal-2022-noisy}, ~\citet{le-etal-2022-shot}, LARA~\cite{huang2025divide}, and \textbf{LENS}~\citep{li2023finding} were adjusted in terms of tasks and models to fit our experimental setup.
We also report the results of fine-tuning the target model using %a parameter-efficient fine-tuning method, namely 
LoRA~\citep{hu2022lora}, %this is a strong baseline that 
which requires access to the model weights. 
Finally, we study \textbf{\moicl Uniform} (\emph{uniform}), an ablation that weights all experts equally, \ie $\forall i: w_{i} = 1/k$.
More implementation details of the baselines can be found in \cref{appendix:imple_details}.

\paragraph{Evaluation Metrics}
For classification tasks, we use accuracy as the evaluation metric.
We use EM (Exact Match) for NQ-open and accuracy for GSM8K.
For preference fine-tuning (\cref{appendix:DPO}), we measure accuracy using the log probabilities of chosen and rejected responses.

More detailed settings, including hyperparameters (\cref{appendix:hyperparameters}) and implementation details with the training process (\cref{appendix:imple_details}), are provided in \cref{appendix:experiment_settings}. Furthermore, in \cref{ssec:partition}, we show that our method is not significantly affected by the choice of partitioning methods. Therefore, we applied static partitioning in all experiments.

\paragraph{Research Questions}
In our experiments, we aim to answer the following questions:
\begin{inparaenum}[(1)]
\item Does \moicl demonstrate general performance improvements over concat-based ICL and other baselines? 
(\S \ref{ssec:classification_tasks} and \S \ref{ssec:partitioning_size})
%(\cref{ssec:classification_tasks,ssec:partitioning_size})
%
\item Does enforcing the mixture-weights to be non-negative improve the downstream accuracy of \moicl? 
%(\cref{sec:anti_expert})
(\S \ref{sec:anti_expert})
\item Can \moicl be extended across multiple language models? 
%(\cref{ssec:moicl_proxy})
(\S \ref{ssec:moicl_proxy})
\item Is MoICL resilient to problem settings involving label imbalance and perturbation? 
%(\cref{sec:OOD,sec:Imbalance,sec:noised_demon})
(\S \ref{sec:OOD} to \S \ref{sec:noised_demon})
\item Can we select demonstrations (experts) based on the tuned weights? 
%(\cref{ssec:select_demon})
(\S \ref{ssec:select_demon})
\item Can \moicl handle demonstrations that were not seen during fine-tuning? 
%(\cref{ssec:hypernet})
(\S \ref{ssec:hypernet})
\item Is \moicl more data, time, and memory-efficient compared to traditional concat-based ICL? %(\cref{ssec:data_eff})
(\S \ref{ssec:data_eff})
%
%\item Can we perform preference fine-tuning in an ICL setting using MoICL?
%
\end{inparaenum}

\subsection{\moicl in Classification Tasks}
\label{ssec:classification_tasks}
To determine the effectiveness of \moicl across various datasets, we compare it with baseline methods in ~\cref{table:various_dataset}. In this experiment, we set the total number of demonstrations ($n$) as 30 and the number of subsets ($k$) as 5, 10, and 30. \moicl outperformed the Baseline ICL on the Offensive, Hate, FEVER, PAWS, and QNLI datasets. The exceptions are SST2 and RTE, where \moicl performs similarly to concat-based ICL in SST2 and shows lower performance in RTE.
The exception in RTE may be attributed to the inherent characteristics of the dataset; as noted in prior works~\cite{dodge2020fine,mosbach2021on,du-nguyen-2023-measuring}, RTE is known to be particularly unstable, which may be further exacerbated in the context of initial demonstration selection.
Surprisingly, \moicl \emph{scalar} achieved the highest performance with $k$=10 (\eg in Hate \moicl achieves 66.52, which is about 10 points increase compared to the concat-based ICl) or $k$=30 (\eg in Offensive \moicl achieves 81.33), rather than $k$=5, in all tasks except for SST2 and RTE. Considering that a larger $k$ reduces the context length (which will be further discussed in~\S \ref{ssec:data_eff}), \moicl manages to capture both efficiency and effectiveness.

\subsection{Impact of Partitioning Size}
\label{ssec:partitioning_size}

\begin{figure}[t]
    \centering
    \includegraphics[width=0.99\columnwidth]{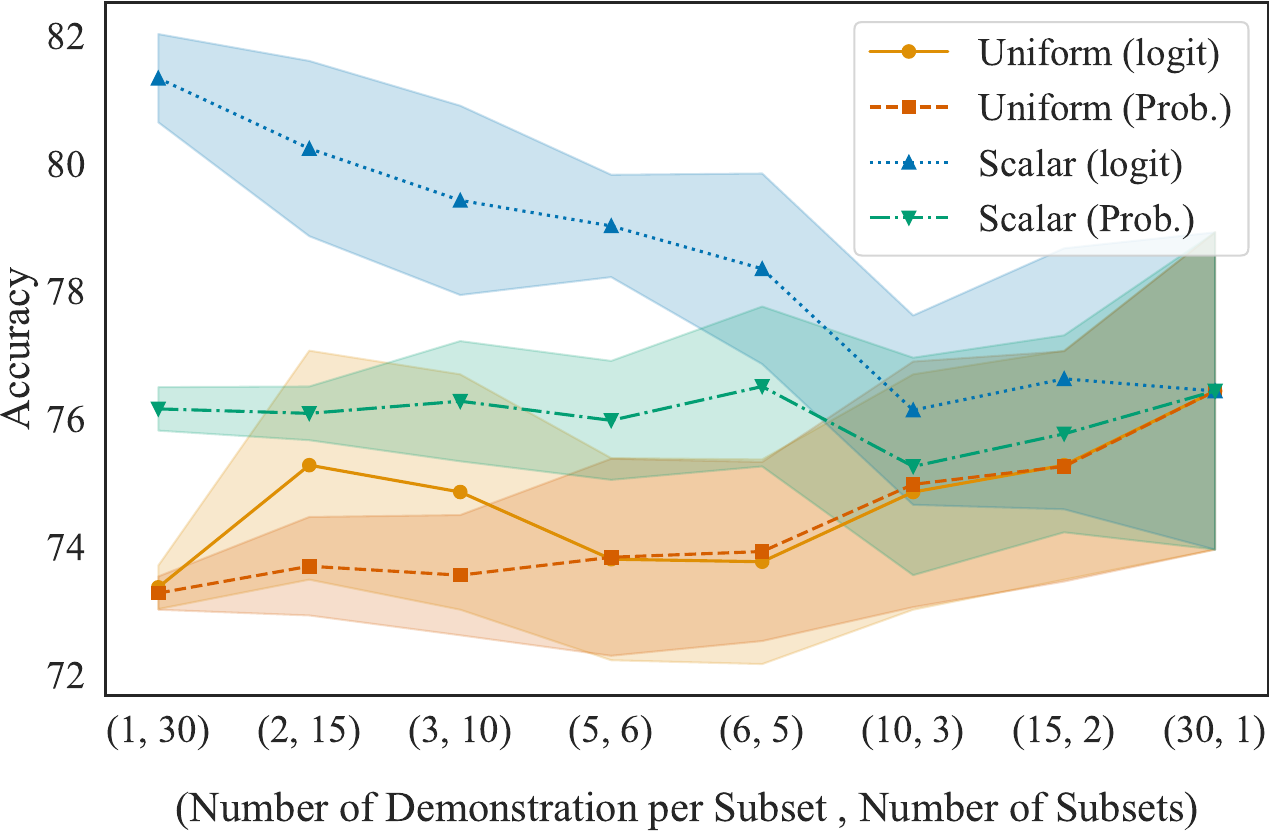}
    \caption{Test accuracy according to the number of demonstrations per subset on TweetEval Offensive dataset. The shaded area represents one standard deviation. We also compare mixing logits to mixing probabilities; see~\cref{appendix:mixing}.
    }
    %\vspace{-0.2cm}
    \label{fig:figure3}
\end{figure}

In ~\cref{fig:figure3}, we present the performance changes on the test set of TweetEval offensive when varying the number of subsets, $k$. 
Since the total number of demonstrations is fixed at 30, each subset contains $30/k$ demonstrations, which corresponds to the x-axis of the Figure. Note that when the number of demonstrations per subset is 30 ($k=1$), it corresponds to the standard Concat-based ICL. We observe that \emph{Uniform Weights} and \emph{scalar} exhibit distinctly different patterns. With \emph{Uniform Weights}, as the number of demonstrations per subset decreases, performance tends to decline, which is an expected outcome for ICL. 
However, with \emph{scalar}, performance surprisingly increases. This seems to be because the decrease in the number of demonstrations per subset is outweighed by the increased flexibility afforded by having more subsets, each assigned tuned weights by \emph{scalar}.

\subsection{Impact of Non-Negative Mixture Weights} \label{sec:anti_expert}
\begin{table}[t]
%\vspace*{-1em}
\centering
\resizebox{0.85\linewidth}{!}{
\begin{tabular}{@{}lc}
\toprule
\textbf{\moicl Method ($n=k=30$)}               & \textbf{Accuracy}               \\
\midrule
\emph{uniform} & 76.44\std{2.48}                  \\
\emph{scalar} ($w \in \mathbb{R}$) & \textbf{81.33\std{0.69}}               \\
\emph{scalar} ($w \in \mathbb{R}_{+}$) (ReLU) & 76.05\std{0.55}\\
\emph{scalar} ($w \in \mathbb{R}_{+}$) (Softplus) & 73.70\std{0.60}\\
\emph{scalar} ($w \in \mathbb{R}_{+}$) (Positive Experts) & 74.47\std{0.68}\\
\bottomrule
\end{tabular}
}
\caption{How important is it to be able to detect anti-experts? 
Results on the TweetEval Offensive Test set using Llama-3-8B-Instruct. 
``\emph{scalar} $w \in \mathbb{R}_{+}$'' constraints the weights $w$ to non-negative values. %, preventing subsets from acting as anti-experts. 
$n$ and $k$ refer to the number of demonstrations and the number of subsets.
}
\label{table:anti-expert}
\end{table}
In Mixtures of Experts, an \emph{anti-expert} is a model that performs poorly on some inputs.
Inspired by \citet{liu2024tuning}, in \moicl, we assume that each expert could also serve as an anti-expert by allowing the expert weights to be negative; if the weight $w_{i} \in \mathbb{R}$ is negative, this indicates that the corresponding expert is actively being used as an anti-expert in generating the response.
To assess the utility of anti-experts in \moicl, we conducted an ablation study in which expert weights were constrained to be non-negative. Specifically, we applied a ReLU activation function ~\cite{agarap2018deep} and a Softplus function (to constrain weights in a more gradient-friendly manner) during training to ensure that all weights remained non-negative, and the same constraint was enforced at inference time. We also implement \textbf{Positive Experts} method that uses weights learned from the standard \emph{scalar} setting but ignores any expert with a negative weight during prediction.

As shown in \cref{table:anti-expert}, this restriction significantly degraded performance (81.33 $\rightarrow$ 76.05 with ReLU), suggesting that anti-experts play a critical role in improving prediction quality. Among the non-negative methods, using ReLU led to better performance than Softplus, indicating that enforcing a hard zero cutoff is more effective than a smooth constraint. Furthermore, the \textbf{Positive Experts} strategy also resulted in a notable performance drop. These findings collectively highlight the importance of allowing the model to leverage both positively and negatively contributing experts. By doing so, \moicl can not only promote useful subsets but also actively suppress useless or misleading ones, ultimately leading to more robust and accurate predictions.

\subsection{\moicl with Multiple Models} \label{ssec:moicl_proxy}

\begin{table}[t]
%\vspace*{-1em}
\centering
\resizebox{0.75\linewidth}{!}{
\begin{tabular}{@{}lc}
\toprule
\textbf{Method ($n=6$)}               & \textbf{GSM8K (Acc.)}               \\
\midrule
%\multicolumn{2}{c}{\textit{Concat-based ICL}}\\
%\midrule
Llama-3.2-3B & 24.76\std{3.12} \\
Llama-3.2-1B-Instruct & 26.47\std{4.27} \\
Llama-3.2-1B & 6.34\std{0.91} \\
\midrule
% \multicolumn{2}{c}{\textit{\moicl (scalar, $k=3$)}}\\
% \midrule
% - Llama-3.2-3B & 21.44\std{2.15} \\
% - Llama-3.2-1B-Instruct & 21.73\std{1.91} \\
% - Llama-3.2-1B & 4.03\std{0.77} \\
% \midrule
\multicolumn{2}{c}{\bf{Multiple models}}\\
\midrule
Proxy-Tuning \cite{liu2024tuning} & 33.63\std{3.99} \\
\midrule
\moicl (\emph{uniform}) & 24.66\std{1.12} \\
\moicl (\emph{scalar}) & 28.93\std{1.62} \\
\moicl (\hypernet) & \textbf{35.65\std{1.45}}\\
\bottomrule
\end{tabular}
}
\caption{Comparison of Concat-based ICL, Proxy-Tuning, and \moicl with multiple models (\S\ref{ssec:moicl_proxy}) on the GSM8K dataset using Llama-3.2-1B, Llama-3.2-1B-Instruct, and Llama-3.2-3B-Instruct. $k$ represents the number of demonstrations subset, where the total number of demonstrations is $n$.}
%
%\EK{For many experiments, hyper-network isn't used. why is it used here?}}
%
\label{table:proxy_tuning}
\end{table}

%
%In \moicl, it is possible for the in-context experts introduced in \cref{sec:moicl} to use different base LLMs.
%
%\EK{I'm missing a bit the motivation of this experiment. When would you want to do this?}
%
For efficiency reasons, the experts in \moicl (\S \ref{sec:moicl}) use the same base model; however, this is not a strict requirement, and experts can also use different base models.\footnote{When vocabularies do not match, techniques like Twist Decoding~\citep{DBLP:conf/emnlp/KasaiS0PLRCS22} can be used.}
To analyse the impact of using \moicl with several models, we draw inspiration by Proxy-Tuning~\cite{liu2024tuning} and consider a \moicl composed of three models: two smaller models (Llama-3.2-1B and Llama-3.2-1B-Instruct, respectively), and a larger base model (Llama-3.2-3B) and evaluate it on GSM8K.
Among the baselines, we also consider the recently proposed Proxy-Tuning~\cite{liu2024tuning}, which merges the next-token distribution of multiple models.
%

%\cref{table:proxy_tuning} compares concat-based ICL, Proxy-Tuning setting (where weights for the large base model, base model, and instruction-tuned model are [1.0, -1.0, 1.0], respectively) and \moicl applied to multiple models.
%
%While \moicl applied to each model individually resulted in lower accuracy compared to concat-based ICL,
%
Results are outlined in \cref{table:proxy_tuning}.
\moicl with the three models led to an improvement of approximately 9.18\% from concat-based ICL and 1.3\% from Proxy-Tuning.
Additionally, an examination of the learned weights for each model revealed values of [\text{0.52\std{0.02}}, \text{-0.21\std{0.03}}, \text{0.69\std{0.02}}] for the large base model, base model, and instruction-tuned model, respectively—aligning with the insights and findings from Proxy-tuning.
This demonstrates that \moicl is extensible to multiple models and serves as a generalization of Proxy-Tuning.

\begin{table}[t]
\centering
\resizebox{1\linewidth}{!}{
\begin{tabular}{@{}lccc}
\toprule
\textbf{Method ($n=k=30$)}               & $p=0.0$ & $p=0.5$ & $p=0.7$ \\
\midrule
Concat-based ICL                  & 76.44\std{2.48}   & 70.67\std{5.06}   & 68.49\std{4.34}   \\
Mixture of ICL & & & \\
\quad\emph{uniform} & 73.37\std{0.34}   & 72.07\std{0.38}  & 70.79\std{0.56}   \\
\quad\emph{scalar} & \textbf{81.33\std{0.69}}   & \textbf{80.95\std{0.65}}   & \textbf{80.19\std{0.37}}  \\
\bottomrule
\end{tabular}
}
\caption{Analysis of out-of-domain (OOD) demonstrations on TweetEval offensive test set using Llama-3-8B-Instruct. Here, $p$ represents the proportion of OOD demonstrations sampled from the SST2 dataset. $n$ and $k$ refer to the number of demonstrations and subsets.
%Bold text signifies the highest accuracy for each value of $p$.
}
\label{table:ood}
\end{table}

\subsection{Handling Out-of-domain Demonstrations}
\label{sec:OOD}
By learning to associate a weight to each expert, \moicl can be used to identify whether demonstrations are relevant to the task.
To analyse this, in \cref{table:ood}, we present the accuracy of \moicl on the TweetEval offensive test set, using a mix of demonstrations sampled from the SST dataset and those from the TweetEval offensive dataset.
We observe that as $p$ (the proportion of OOD demonstrations) increases, the performance of standard ICL methods decreases. However, \moicl (with \emph{scalar}) effectively mitigates this by reducing the influence of these OOD demonstrations, resulting in the smallest performance drop. 

This becomes even more apparent when analysing the weights of actual OOD demonstrations. 
When $p=0.5$ (\ie the number of OOD and in-domain demonstrations is equal), the average weight of in-domain demonstrations is \text{0.0108\std{0.0025}}, while the average weight for OOD demonstrations is \text{-0.0059\std{0.0027}}. For $p=0.7$, the average weight of in-domain demonstrations is \text{0.0127\std{0.0052}}, while the average weight for OOD demonstrations is \text{-0.0019\std{0.0016}}.
\begin{figure}[t]
    \centering
    \begin{subfigure}[b]{0.49\linewidth}
        \centering
        \includegraphics[width=\linewidth]{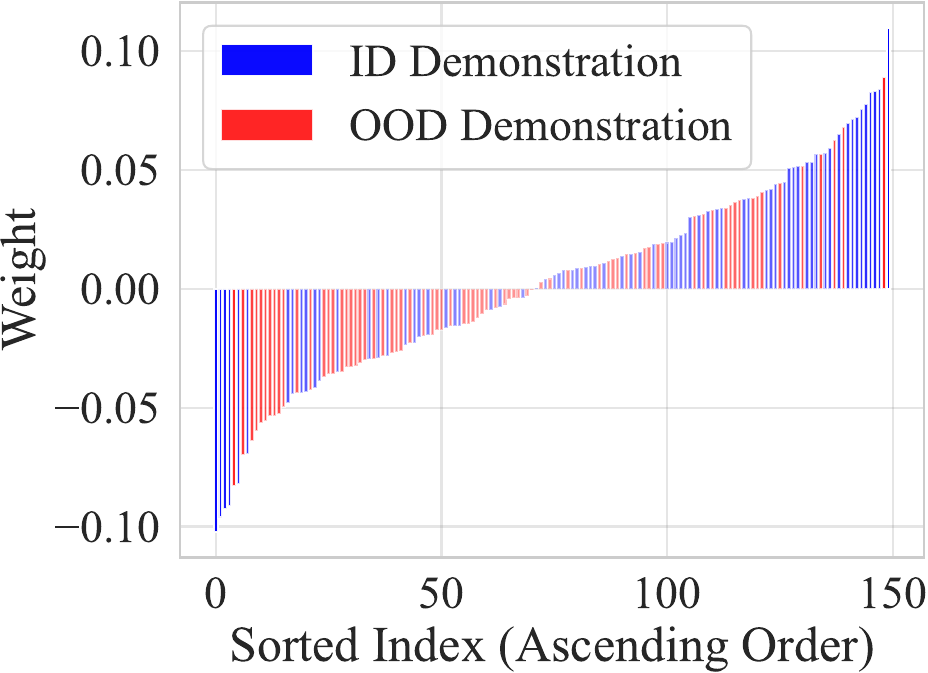}
        \caption{50\% OOD}
    \end{subfigure}
    \begin{subfigure}[b]{0.49\linewidth}
        \centering
        \includegraphics[width=\linewidth]{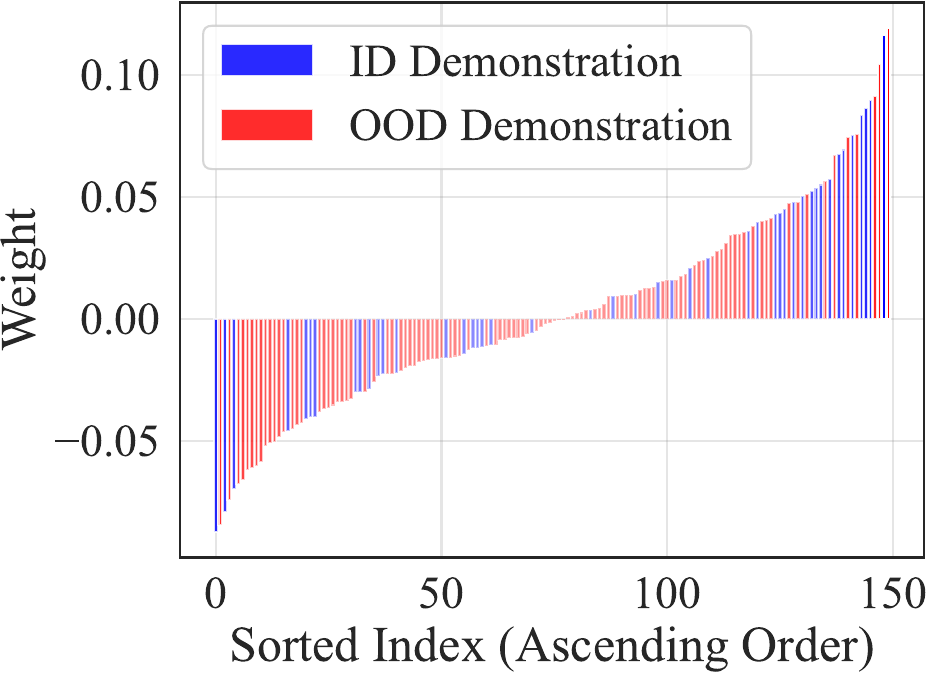}
        \caption{70\% OOD}
    \end{subfigure}
\caption{Visualisation of the tuned weights when (a) 50\% and (b) 70\% of demonstrations are OOD. The y-axis indicates the weights, whereas the x-axis represents the index of demonstrations sorted in ascending order (across five different seeds). Blue bars correspond to in-domain (ID) demonstrations, and red bars correspond to out-of-domain (OOD) demonstrations. 
}
\label{fig:ood_visual}
\end{figure}
In ~\cref{fig:ood_visual}, we show how the weights of in-domain demonstrations (blue bars) and OOD demonstrations (red bars) are distributed.
We can see that in-domain demonstrations typically receive positive weights, while OOD demonstrations tend to receive negative weights.
This provides evidence that \moicl successfully mitigates the influence of OOD demonstrations.

\subsection{Mitigating Label Imbalance}
\label{sec:Imbalance}

\begin{table}[t]
\centering
\resizebox{0.85\linewidth}{!}
{
\begin{tabular}{@{}lcc}
\toprule
\textbf{Method ($n=k=30$)}               & \textbf{Original} & \textbf{Imbalanced} \\
\midrule
Concat-based ICL                  & 76.44\std{2.48}      & 28.49\std{0.86}   \\
Mixture of ICL & & \\
\quad\emph{uniform} & 73.37\std{0.34}    & 40.19\std{2.32}     \\
\quad\emph{scalar} & \textbf{81.33\std{0.69}}     & \textbf{77.77\std{1.20}}    \\
\bottomrule
\end{tabular}
}
\caption{Analysis of imbalanced demonstrations on the TweetEval Offensive Test set using Llama-3-8B-Instruct. ``Imbalanced'' refers to a condition where only one out of 30 demonstrations has a ``neutral'' label, while the rest are ``offensive''.}
%\vspace*{-1em}
\label{table:imbalanced}
\end{table}

To determine whether our proposed method can handle label imbalance, on the TweetEval Offensive dataset, we set up 29 ``offensive'' label demonstrations and one `non-offensive' label demonstration out of 30 demonstrations. Since the TweetEval Offensive dataset has a ``non-offensive'' to ``offensive'' label ratio of 7:3, such imbalanced demonstrations would be detrimental to performance. As seen in \cref{table:imbalanced}, such imbalanced demonstrations caused a significant performance drop in standard ICL methods. However, \moicl (\emph{scalar}) showed the least performance drop by mitigating the effects of label imbalance.

\subsection{Filtering Perturbed Demonstrations}
\label{sec:noised_demon}

\begin{figure}[t]
    \centering
    \includegraphics[width=\linewidth]{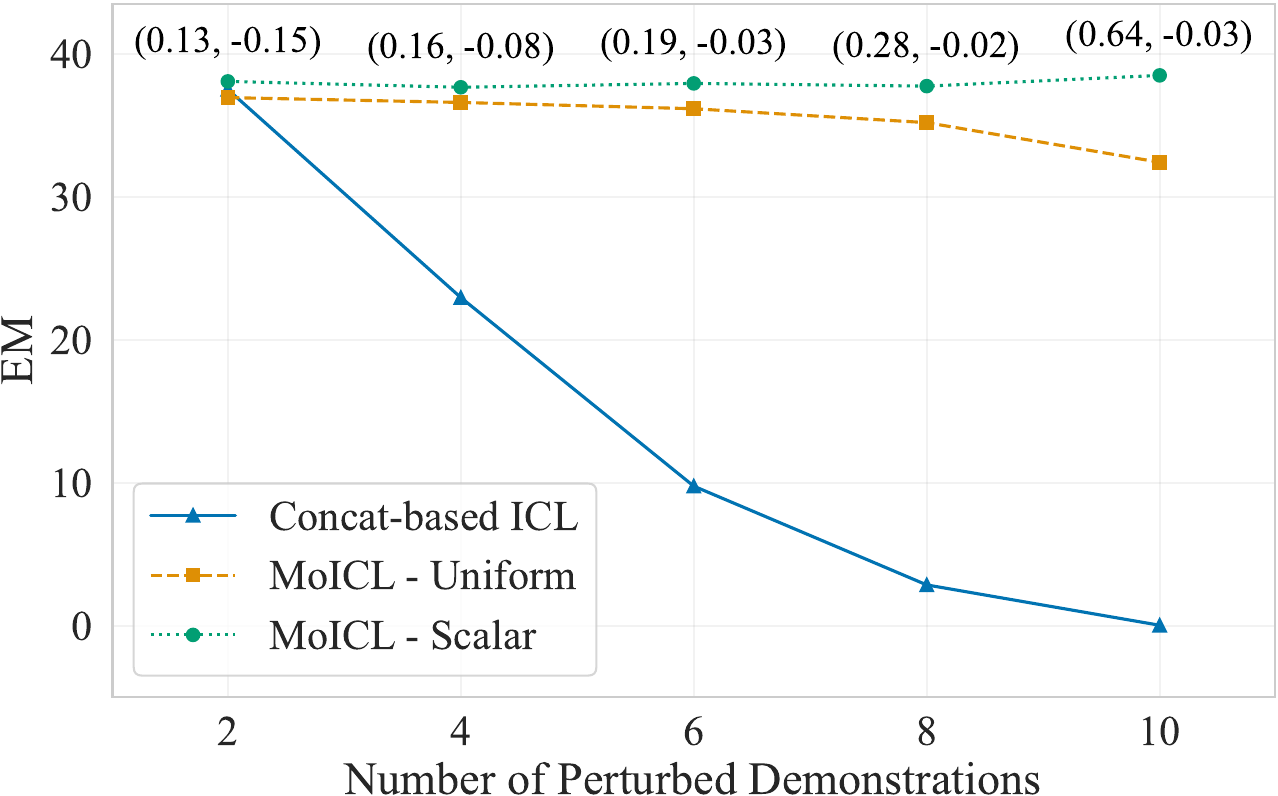}
    \caption{Resilience of ICL to adding perturbed demonstration. We report the EM based on the number of perturbed demonstrations out of the total 12 demonstrations in NQ. For the case of \emph{scalar}, we also present the average weights of standard and perturbed demonstrations as (standard, perturbed).
    }
    \label{fig:figure_noise}
\end{figure}

\begin{table*}[ht]
\small
\centering
\begin{tabular}{@{}lccccc}
\toprule
\textbf{Method $\downarrow$ Subset $\rightarrow$} & \textbf{$k^{\prime}=5$} & \textbf{$k^{\prime}=10$} & \textbf{$k^{\prime}=20$} & \textbf{$k^{\prime}=30$} & \textbf{$k^{\prime}=90$} \\
\midrule
Concat-based ICL ($n=k^{\prime}$)        & 72.19\std{2.63} & 74.12\std{2.24}  & 74.84\std{1.88}  & 76.44\std{2.48} & 75.67\std{2.33}  \\\midrule
\moicl ($n, k = k^{\prime}$)         &       &       &       &       &       \\
\quad\emph{uniform}         & 73.05\std{0.52}  & 73.42\std{0.76}  & 73.42\std{0.49} & 73.37\std{0.34} & 73.26\std{0.16} \\
\quad\emph{scalar}          & \textbf{76.26\std{1.11}} & \textbf{78.16\std{0.91}} & \textbf{80.16\std{1.23}} & \textbf{81.33\std{0.69}} & \textbf{83.35\std{0.41}} \\
\midrule
\moicl \emph{scalar} ($n=k=90$)       &  &  &  &  &      \\
\quad Highest $k^{\prime}$ Weights       & 75.58\std{0.81}  & 75.56\std{0.46} & 74.42\std{0.61} & 74.33\std{0.38} & -     \\
\quad Highest $k^{\prime}$ Weights (abs) & 69.79\std{14.84} & 60.53\std{21.84} & 71.58\std{14.67} & 72.93\std{13.14} & -     \\
\quad IMLE Top-$k^{\prime}$ mask%\citep{niepert2021implicit}    
& \textbf{76.07\std{0.64}} & \textbf{75.93\std{0.69}} & \textbf{76.35\std{0.35}} & \textbf{76.44\std{0.64}} & -    \\
\bottomrule
\end{tabular}
\caption{Analysis of selecting useful demonstrations with the proposed \moicl on the TweetEval Offensive test set on Llama-3-8b-Instruct. `Highest $k^{\prime}$ Weights' refers to selecting the $k^{\prime}$ subsets with the largest weights out of 90 weights of \moicl \emph{scalar}, while `Highest $k^{\prime}$ Weights (abs)' uses absolute weights instead. 
}
\label{table:select}
\end{table*}

\begin{table*}[t]
\centering
\resizebox{\linewidth}{!}{
\begin{tabular}{@{}lccccccc}
\toprule
\textbf{Method $\downarrow$ Dataset $\rightarrow$}               & \textbf{Offensive} & \textbf{Hate} & \textbf{SST} & \textbf{RTE} & \textbf{FEVER} & \textbf{PAWS} & \textbf{QNLI}\\
\midrule
Concat-based ICL ($n=30$)   & 76.44\std{2.48}       & 53.54\std{4.29}  & \textbf{95.46\std{0.14}}  & \textbf{86.43\std{1.26}} &   80.63\std{0.49}  &   \textbf{78.12\std{0.77}}  & 89.08\std{0.44}\\
%\midrule
Mixture of ICL ($n=k=30$) &                   &\\
\quad\emph{uniform} & 73.37\std{0.34}       & 59.12\std{0.47}  & 94.17\std{0.21}  & 77.26\std{1.02} &  79.46\std{0.36} &  65.29\std{0.51} & 88.66\std{0.25}\\
\quad\hypernet    &  \textbf{77.33\std{3.28}} & \textbf{65.01\std{4.10}} & 94.24\std{0.50} & 80.00\std{0.59} & \textbf{80.68\std{0.32}} & 73.79\std{0.14} & \textbf{89.21\std{0.26}}\\
\bottomrule
\end{tabular}
}
\caption{Comparison of \moicl methods on unseen demonstrations, including \emph{uniform} and \hypernet, on the TweetEval Offensive and Hate, using Llama-3-8b-Instruct. 
}
\label{table:hypernet}
\end{table*}

One of the benefits of assigning weights to each demonstration or its subsets is the ability to handle low-quality, or more specifically, perturbed demonstrations. To verify this, in NQ-Open, we perturbed demonstrations (see \cref{appendix:genration} for the result of NQ-open without perturbation) by randomly changing the answers to one of (\textit{yes}, \textit{no}, \textit{foo}, \textit{bar}), where the total number of demonstration is 12, and each subset has one demonstration ($n$, $k=12$). 
The results in \cref{fig:figure_noise} show that our proposed method effectively handles perturbed demonstrations.
While the EM of the concat-based ICL significantly decreases as the number of perturbed demonstrations increases, the \moicl methods can maintain performance.
Additionally, without tuning the weights (\emph{uniform}), performance gradually declines as the number of perturbed demonstrations increases, but with tuning (\emph{scalar}), the performance remains stable (more than +35\% with 10 perturbed demonstrations).
This is clear when analysing the weights; \cref{fig:figure_noise} shows the average weights of normal and perturbed demonstrations, with significantly lower weights for the latter (\eg, 0.64 vs -0.03). %, showing a noticeable difference. \pmm{can we say what we see in the weights? let's assume the reviewer is lazy}

\subsection{Selecting Demonstration Subsets}
\label{ssec:select_demon}
We now analyse the impact of sparsifying the mixture weights $w \in \mathbb{R}^{k}$ in \moicl.
Results are available in \cref{table:select} --- ``Highest $n$ Weights'' refers to selecting the subsets with the $n$ largest $w$ weights (or $|w|$ in the case of ``abs''), while IMLE Top-$k^{\prime}$ mask refers to the method introduced in \S \ref{sec:moicl}, %using $\lambda = 1$ 
following the default hyper-parameters proposed by \citet{niepert2021implicit}.
While \moicl (\emph{scalar}) achieved the highest accuracy, the need to learn them for each $m$ and $k$ makes selection methods that tune weights for a large $n$ and then select $m$ of them more practical.
Notably, ``Highest $n$ Weights (abs)'' is high-variance, indicating the difficulty in effectively leveraging anti-experts (\S \ref{sec:anti_expert}).
In contrast, IMLE, which uses a mask, demonstrated stable performance, achieving the best results even with a limited number of demonstrations. %, particularly with a few demonstrations (when $k^{\prime}=5$).
This also suggests that \moicl can be interpreted through the lens of a mixture-of-experts paradigm: rather than relying solely on the highest-weighted experts, performance gains emerge from expert coordination and interaction among experts within a broader set.

\subsection{Generalisation to Unseen Demonstrations}
\label{ssec:hypernet}
While \moicl in the \emph{scalar} setting is conceptually simpler, it has the disadvantage of requiring a fixed set of demonstration subsets.
A solution to overcome this limitation is to utilise a smaller, fine-tuned hyper-network (\hypernet) that calculates the weights for arbitrary demonstration subsets.
\cref{table:hypernet} compares the performance of \moicl with different choices for the mixing weights in settings where the demonstration set $D$ was not available during the training process.
In this situation, \emph{scalar}, which assumes that experts and their demonstrations are available beforehand, cannot be used.
However, a hyper-network trained to map sets of demonstrations to mixture weights can generalise even when presented with unseen demonstrations.

Compared to the \emph{uniform} baseline, \moicl consistently outperforms it, indicating that the hyper-network can assign meaningful, appropriate weights to arbitrary demonstration subsets. When compared to the concat-based ICL baseline, \moicl achieves competitive or superior performance on 4 out of 7 tasks. Considering that the concat-based ICL baseline uses full $n$-shot demonstrations while the adaptive variant of \moicl effectively operates in a much lower $(n/k)$-shot setting, this level of performance can be considered reasonable.

\section{Data and Compute Efficiency} \label{ssec:data_eff}

\begin{figure}[t]
    \centering

    \begin{subcaptionbox}{TweetEval Offensive\label{fig:figure_data_eff_off}}{
        \includegraphics[width=\linewidth]{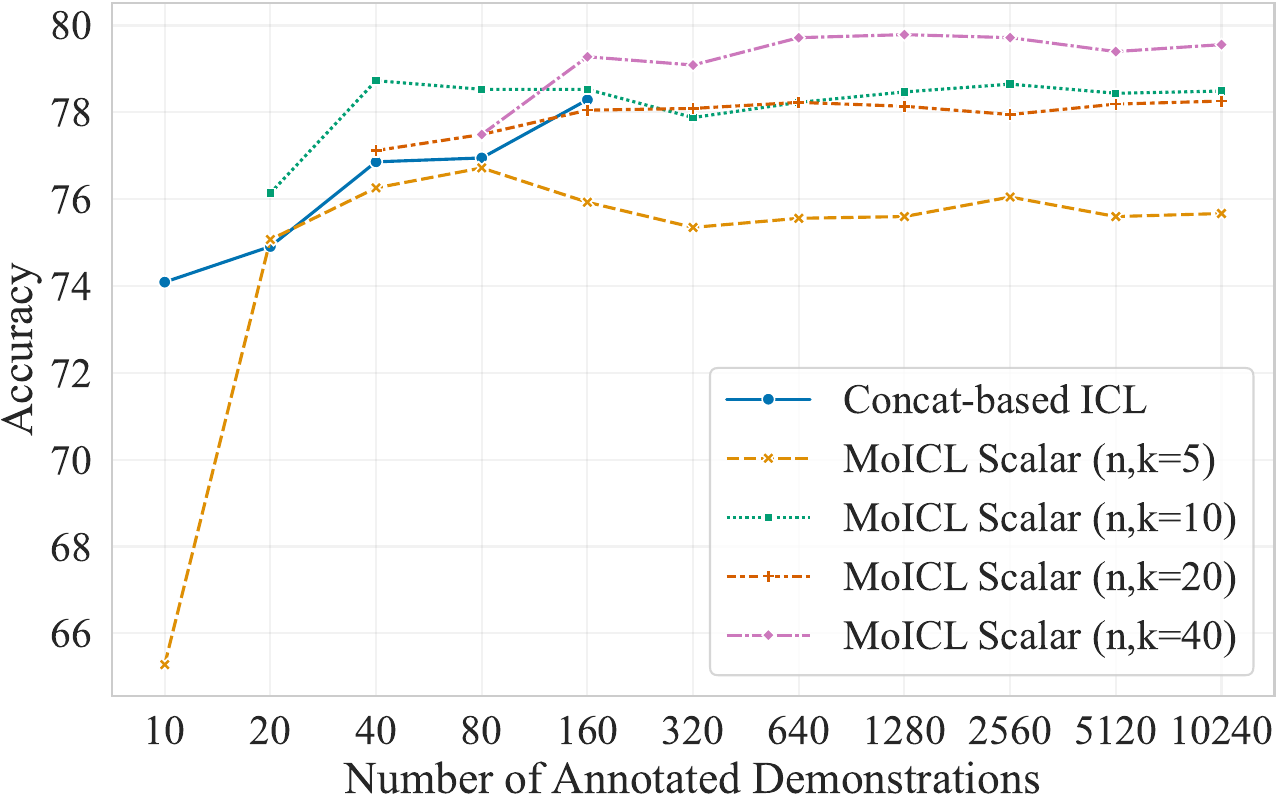}
    }
    \end{subcaptionbox}

    \begin{subcaptionbox}{TweetEval Hate\label{fig:figure_data_eff_hate}}{
        \includegraphics[width=\linewidth]{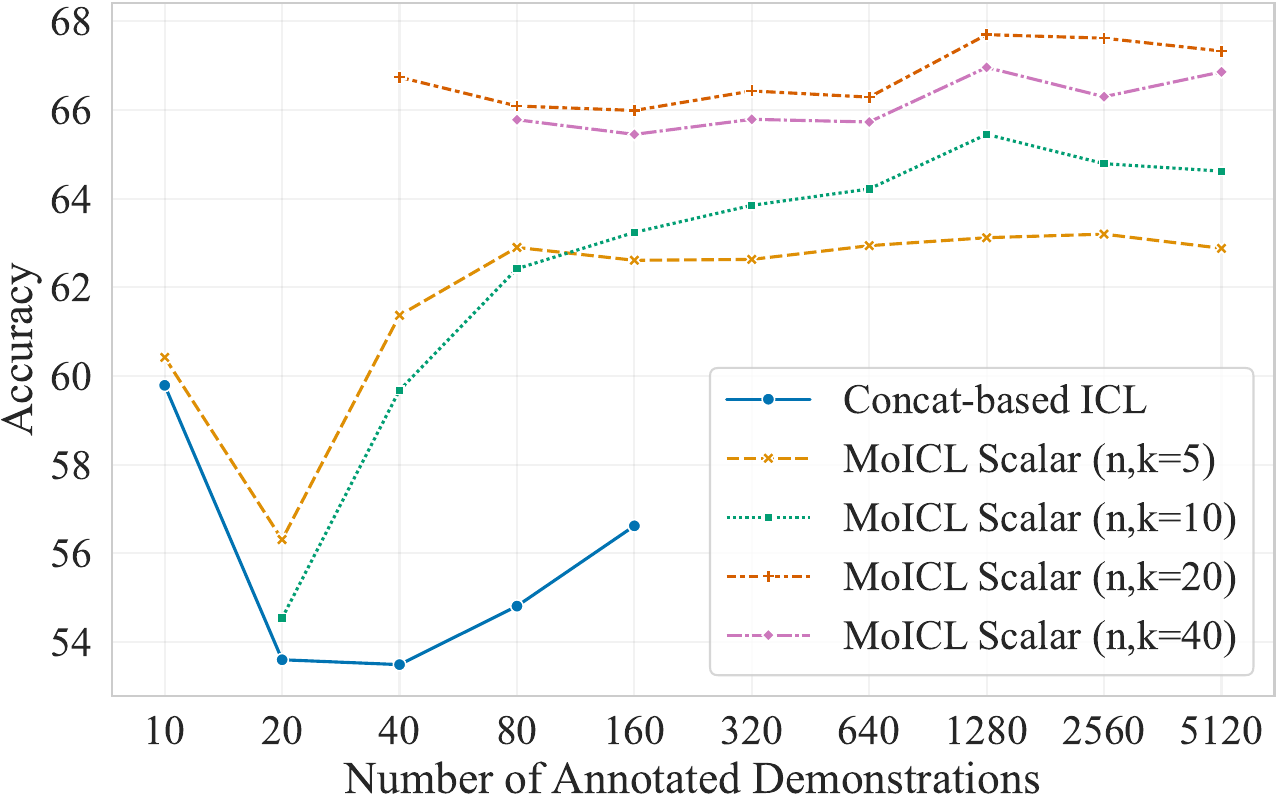}
    }
    \end{subcaptionbox}
    
    \caption{An analysis of \moicl's data-efficiency on the TweetEval offensive/hate test set using Llama-3-8B-Instruct. Concat-based ICL concatenated all available demonstrations ($x$-axis), exceeding the context length when ($n>160$).}
    %
    %\moicl \emph{scalar} ($k=n$) assigned the designated demonstrations to the experts while using the remaining available demonstrations for fine-tuning.}
    \label{fig:figure_data_eff}
\end{figure}
%
%One potential limitation of \moicl is that it requires training instances for weight tuning, which can be problematic when such training data is unavailable.
%
To analyse the data-efficiency of \moicl, we present the accuracy on TweetEval Offensive and Hate test set in ~\cref{fig:figure_data_eff} under settings where the number of training instances (Number of Annotated Demonstrations) is limited. In this experiment, we set $n=k$, so each expert is assigned one demonstration and weight tuning is performed using the number of training instances minus $k$ (e.g., when the x-axis is at 40, \moicl with $k=10$ is tuned with 30 training instances). We observed that \moicl is highly data-efficient, achieving better performance than concat-based ICL with only around 20 annotated demonstrations. In contrast, concat-based ICL showed lower performance when given the same number of annotated demonstrations and particularly struggled when the number of demonstrations exceeded 160, exceeding the maximum allowed context length.

\begin{figure}[t]
    \centering
    
    \begin{subcaptionbox}{TweetEval Offensive\label{fig:figure_time_eff_off}}{
        \includegraphics[width=\linewidth]{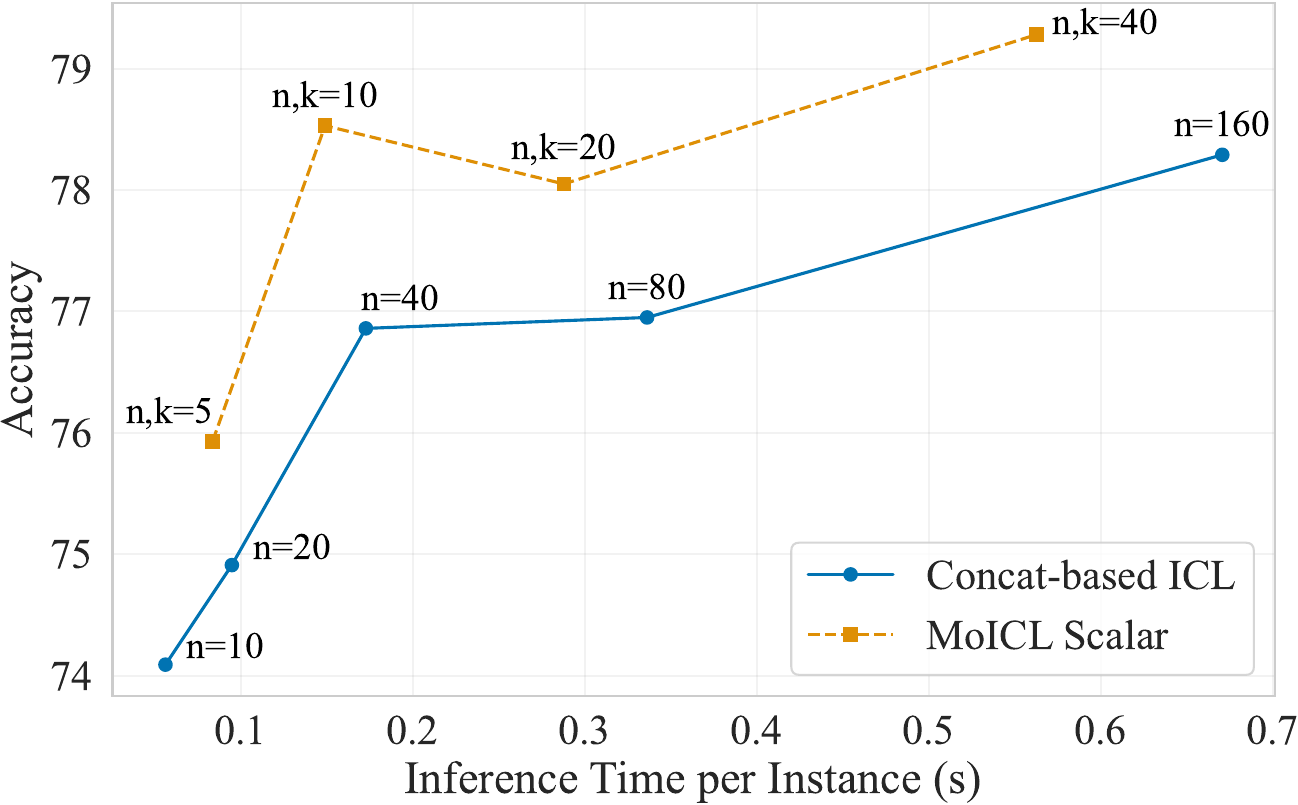}
    }
    \end{subcaptionbox}

    \begin{subcaptionbox}{TweetEval Hate\label{fig:figure_time_eff_hate}}{
        \includegraphics[width=\linewidth]{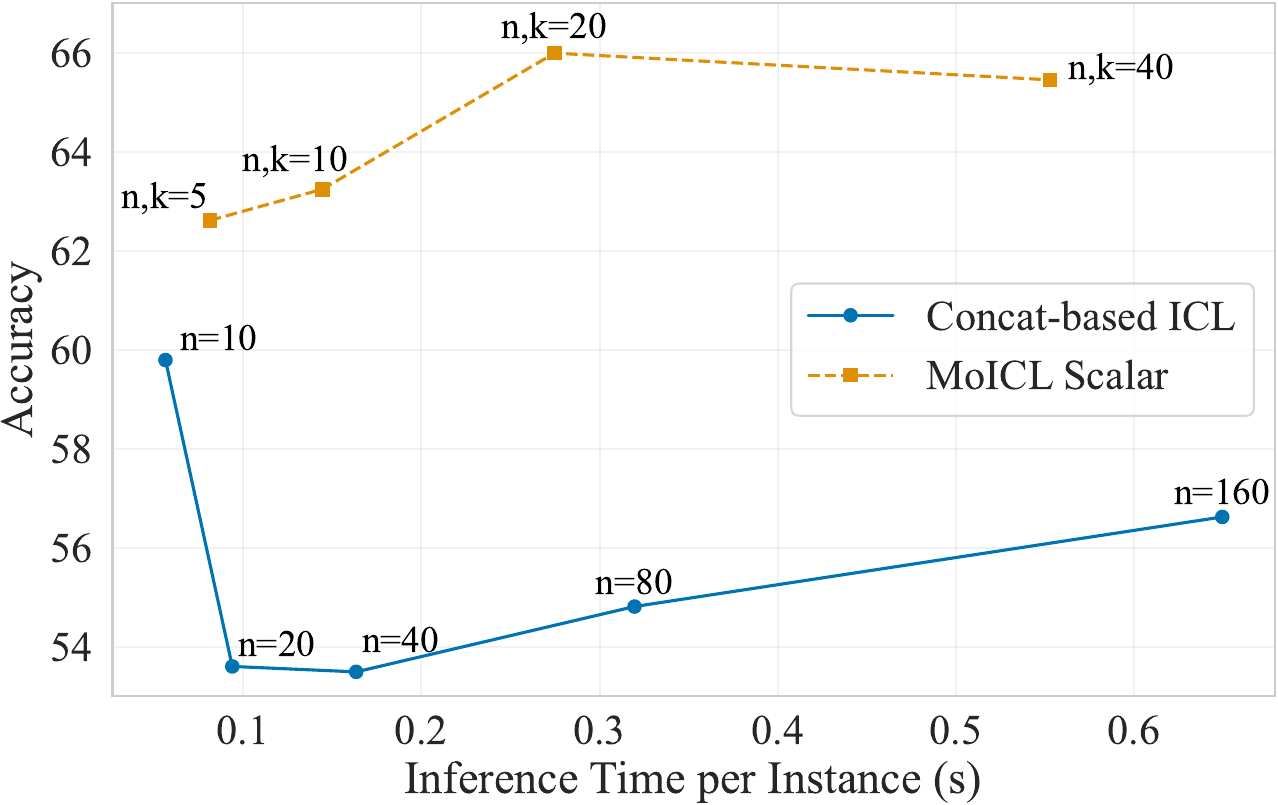}
    }
    \end{subcaptionbox}
    \vspace*{-1em}
    \caption{An analysis of the inference time efficiency of \moicl on the TweetEval offensive/hate test set using Llama-3-8B-Instruct. The total number of demonstrations available to \moicl \emph{scalar} is 160, the same as the maximum number of demonstrations that concat-based ICL can use within the context length limit.}
    \label{fig:figure_time_eff}
\end{figure}

Furthermore, we analyse whether \moicl is more time-efficient than concat-based ICL.
\cref{fig:figure_time_eff} compares the performance in terms of the average inference time (in seconds) per instance with up to 160 annotated demonstrations (which is the context length limit for concat-based ICL) are provided. We observed that \moicl consistently showed higher accuracy compared to concat-based ICL relative to inference time, demonstrating that \moicl is not only data-efficient but also time-efficient. In addition, \cref{appendix:training_cache} shows that \moicl training can also be made more efficient with caching.

\begin{table}[t]
%\vspace*{-1em}
%\vspace*{-1em}
%\small
\centering
\resizebox{\columnwidth}{!}{
\renewcommand{\arraystretch}{1.1}
\begin{tabular}{@{}lc}
\toprule
\textbf{Method}    & \textbf{Complexity} \\
\midrule
Concat-based ICL   &  $(n+1)^2 \cdot C_{\text{LLM}}$                          \\
Ensemble-based ICL &  $n \cdot (1+1)^2 \cdot C_{\text{LLM}}$                         \\
\midrule
Mixture of ICL     &                           \\
\quad \emph{uniform}  &  $k \cdot (\frac{n}{k}+1)^2 \cdot C_{\text{LLM}}$                         \\
\quad \emph{scalar}   &  $k \cdot (\frac{n}{k}+1)^2 \cdot C_{\text{LLM}}$                         \\
\quad \hypernet         &  $k \cdot (\frac{n}{k}+1)^2 \cdot C_{\text{LLM}} + n^2 \cdot C_{\text{Hyper}}$ \\
\bottomrule
\end{tabular}
}
\caption{Comparison of the computational complexity at inference time between \moicl Methods and Baseline ICL Methods. $C_{\text{LLM}}$ and $C_{\text{Hyper}}$ refer to the unit computation complexity for one demonstration and one forward pass for an LLM and a hyper-network, respectively. $n$ and $k$ refer to the number of demonstrations and the number of subsets. 
}
\label{table:complexity}
\end{table}
\paragraph{Complexity} 
The proposed \moicl method partitions demonstrations into subsets rather than concatenating them, thereby reducing the input context length for LLMs.
This reduction is beneficial in Transformer-based architectures, where computational load increases quadratically with the context length.
In \cref{table:complexity}, we analyse the computation cost based on the unit computation cost (one forward pass for one example) of an LLM and a hyper-network, namely $C_{\text{LLM}}$ and $C_{\text{Hyper}}$. 
Concat-based ICL exhibits the highest cost by concatenating all demonstrations and the test input, %$(n+1)$,
whereas Ensemble-based ICL shows the lowest cost by concatenating each demonstration with the test input.
%$(1+1)$.
%
\moicl lies in-between, with the cost determined by the number of subsets $k$.
In \hypernet, the weights calculation adds a cost of $(n + 1)^2 \cdot C_{\text{Hyper}}$.
Since $C_{\text{LLM}}$ is usually significantly larger than $C_{\text{Hyper}}$, this approach still offers a computational advantage.
Furthermore, the weights of the experts need to be computed only once, which means $n^2 \cdot C_{\text{Hyper}}$ is a one-time cost.

\section{Related Work} \label{sec:rw}
\paragraph{In-Context Learning}
In-context learning (ICL) is an approach to few-shot learning by concatenating the training examples and providing them as input to the model before the actual test example.
Being able to perform ICL is an \emph{emerging ability} of very large models, such as GPT-3~\citep{NEURIPS2020_1457c0d6} and PaLM~\citep{DBLP:journals/jmlr/ChowdheryNDBMRBCSGSSTMRBTSPRDHPBAI23}.
One characteristic of ICL is that increasing the number of demonstrations tends to increase the downstream task accuracy~\cite{NEURIPS2020_1457c0d6,lu-etal-2022-fantastically}.
However, \citet{agarwal2024many} show that, after a given number of demonstrations, performance saturates and additional examples might even decrease the downstream task accuracy.
Furthermore, in Transformer-based LLMs, increasing the number of ICL demonstrations can be too computationally demanding due to the complexity of self-attention operations growing quadratically with the context size~\cite{liu2022few}.
Finally, ICL is sensitive to out-of-domain demonstrations~\citep{min-etal-2022-rethinking} or label imbalance, underscoring the importance of the selection of the in-context demonstrations to use~\citep{zhao2021calibrate, fei-etal-2023-mitigating}.

\paragraph{Ensembles of Demonstrations}
%
%\citet{min-etal-2022-noisy} introduce \emph{ensemble-based demonstrations} as an alternative to concat-based ICL (\cref{ssec:ICL}), where each demonstration $(x_i, y_i)$ is provided to a language model along with the input $x$ to obtain a next-token distribution $p\left( y \mid x_i, y_i, x \right)$; such next-token distributions are then combined in a product-of-experts to produce the final next-token distribution: $p\left( y \mid x_{1}, y_{1}, \ldots, x \right) = \prod_i p\left( y \mid x_i, y_i, x \right)$.
%
\citet{min-etal-2022-noisy} introduce \emph{ensemble-based demonstrations}, where each demonstration is provided to a language model along with the input to obtain a next-token distribution; such next-token distributions are then combined in a product-of-experts.
\citet{le-etal-2022-shot} propose Mixtures of In-Context Experts for anaphora resolution, where the weights for each expert were calculated based on the cosine similarity between the embeddings of the test input and the demonstrations.
\citet{ye-etal-2023-fid} extend the models by \citet{le-etal-2022-shot} and analyse the impact of merging the expert activations at different stages, both in terms of efficiency and downstream task performance. \citet{wangmixture} proposed a framework that partitions demonstrations into semantically similar clusters, assigns them to experts, and trains a retriever model for each expert to retrieve demonstrations based on a given input query.
\moicl is related to ensemble-based ICL methods in that it aggregates predictions from multiple demonstration subsets. However, unlike prior approaches that rely on similarity-based retrieval or uniform aggregation, \moicl learns performance-driven weights across fixed or dynamic subsets, enabling more explicit modelling of expert contributions. Most notably, \citet{huang2025divide} (LARA) shares similar motivations and can be considered as a concurrent approach. While LARA uses non-gradient-based optimisation, \moicl leverages gradient-based optimisation with support for supervised fine-tuning and preference tuning (\cref{appendix:DPO}). Additionally, \moicl incorporates a hyper-network (\emph{adaptive}) to generalise to unseen demonstration subsets (\cref{ssec:hypernet}), which addresses the limitation of requiring fixed demonstrations between training and inference. \moicl also introduces the concept of anti-experts in ICL and demonstrates practical effectiveness under challenging conditions such as out-of-domain inputs, label imbalance, and perturbed demonstrations.
%

%Our proposed Mixture of In-Context Learners (\moicl) extends such approaches by learning a \emph{weighting function} assigning specific weights to each expert.
%
%Our experiments show that this approach allows us to tackle various challenges in ICL such as label imbalance, out-of-distribution demonstrations, and sample selection without requiring access to the model weights.
%

\section{Conclusions}
We proposed \moicl, a method for dynamically learning to combine multiple models, each trained via ICL, via gradient-based optimisation methods.
We show that \moicl significantly improves accuracy compared to several strong baselines, as well as in terms of data, memory, and compute efficiency.
Furthermore, we show that \moicl is robust to out-of-domain and perturbed demonstrations, can help mitigate label imbalance, and can be used to select sets of demonstrations.

\section*{Limitations}

Although \moicl does not require direct access to the model parameters, it requires access to the logits of the distribution over the vocabulary or answers produced by the model, both to train the experts and to calculate the final prediction at inference time, which prevents its use with completely black-box models like GPT-4. 
%Future work can consider black-box optimisation methods to address this limitation. 
As discussed and tested in \cref{appendix:black_box}, it is possible to apply \moicl by leveraging an approximation of the full logits in situations where logits are extracted from proprietary, restricted-access models or only top-k tokens are available. However, the requirement for logits themselves (even if only partially) can be a limitation.

An important direction for future work, though not explored in this study, is training scalar weights for all demonstrations in the training set. Currently, we sample $n$ demonstrations from the training set and assign them to experts, tuning their weights. Extending this to all demonstrations in the training set would require progressively expanding the experts and their tuned weights. One possible approach for future work is to incorporate the search and relevance heuristics proposed by \citet{li2023finding} as inductive biases in our proposed \moicl (\hypernet).

Additionally, due to computational resource limitations, we conduct our experiments on the Llama-2 models (Llama-2-7B-chat, Llama-2-13B-chat, Llama-2-70B-chat), Llama-3 models (Llama-3-8B, Llama-3-8B-Instruct), Llama-3.1 models (Llama-3.1-8B, Llama-3.1-8B-Instruct), and Llama-3.2 models (Llama-3.2-3B, Llama-3.2-1B, Llama-3.2-1B-Instruct) as target LLMs, and T5-models (T5-efficient-tiny, T5-efficient-mini, T5-small, T5-base) as hyper-network models (\cref{sec:model_size}). However, our method is not limited to specific LMs and can be applied across various models. 

\paragraph{Acknowledgments}
We thank the anonymous reviewers for their useful feedback and comments.
Giwon Hong was supported by the ILCC PhD program (School of Informatics Funding Package) at the University of Edinburgh, School of Informatics.
Pasquale Minervini and Emile van Krieken were partially funded by ELIAI (The Edinburgh Laboratory for Integrated Artificial Intelligence), EPSRC (grant no.\ EP/W002876/1). Additionally, Pasquale Minervini was partially funded by an industry grant from Cisco and a donation from Accenture LLP.
This work was supported by the Edinburgh International Data Facility (EIDF) and the Data-Driven Innovation Programme at the University of Edinburgh.
%

\bibliography{custom}

\clearpage

\begin{table}[t]
\centering
\resizebox{\linewidth}{!}{
\begin{tabular}{@{}lccccccc}
\toprule
\bf{Split} & \bf{Offensive} & \bf{Hate} & \bf{SST2} & \bf{RTE} & \bf{FEVER} & \bf{PAWS} & \bf{QNLI} \\
\midrule
Train set & 11,916    & 9,000 & 66,349 & 2,190 & 54,550 & 49,401 & 99,743 \\
Dev set   & 1,324     & 1,000 & 1,000  & 300   & 5,000  & 8,000  & 5,000  \\
Test set  & 860       & 2,970 & 872    & 277   & 13,332 & 8,000  & 5463  \\
\bottomrule
\end{tabular}
}
\caption{Statistics of the classification datasets used in our experiments.}
\label{table:stat}
\end{table}

\appendix

\section{Detailed Experiment Settings}
\label{appendix:experiment_settings}

\subsection{Datasets}

TweetEval~\citep{barbieri-etal-2020-tweeteval} offensive/hate datasets are originally from~\citet{zampieri2019semeval} and ~\citet{basile-etal-2019-semeval}, respectively. PAWS~\citep{zhang-etal-2019-paws} is released under a custom license\footnote{The dataset is provided "AS IS" without any warranty, express or implied. Google disclaims all liability for any damages, direct or indirect, resulting from the use of the dataset.} from Google LLC.
For SST-2~\citep{socher-etal-2013-recursive}\footnote{The dataset is released under \texttt{The MIT License}.}, RTE~\citep{bentivogli2009fifth}, FEVER~\citep{thorne2018fever}\footnote{The dataset is released under \texttt{CC BY-SA 3.0} license.}, and QNLI~\citep{wang2018glue}\footnote{The dataset is released under \texttt{CC BY-SA 4.0} license.}, we used the original validation/development set as the test set and sampled a portion of the training set to construct a new validation set. 
~\cref{table:stat} presents the dataset split statistics for all classification datasets used in our experiments. 
For NQ-open~\citep{lee-etal-2019-latent}\footnote{The dataset is released under \texttt{CC BY-SA 3.0} license.}, we used the top 1 retrieved documents as a context. The dataset contains 79,168 train instances, 8,757 validation instances, and 3,610 test instances.
GSM8K~\citep{cobbe2021gsm8k}\footnote{The dataset is released under \texttt{The MIT License}.} contains 7,473 train instances and 1,319 test instances, employing chain-of-thought (CoT).
HH-RLHF~\citep{bai2022training}\footnote{The dataset is released under \texttt{The MIT License}.} Harmless dataset contains 42,537 training instances (chosen, rejected pairs) and 2,312 test instances. 
\subsection{Hyperparameters}
\label{appendix:hyperparameters}
We used five seeds $[31, 42, 65, 438, 991]$ in all experiments except for \cref{appendix:DPO}, which were applied to every possible aspect, including dataset shuffle, demonstration pooling and partition, and hyper-network fine-tuning, and baseline results. 
Also, we set the batch size to 1, the gradient accumulation steps to 12, the warmup steps to 500, the weight decay to 0.01, and the learning rate to 0.0001 without performing a hyperparameter search for these settings. We used AdamW \citep{loshchilov2018decoupled} as an optimiser. We use the same setting for Direct Preference Optimization~\cite[DPO;][]{rafailov2024direct} in \cref{appendix:DPO} as well.
For the PEFT ~\citep[LoRA,][]{hu2022lora} baseline, we set $r$=16 (rank), alpha=32 (scale factor), and dropout=0.1. We did not perform a search for these LoRA hyperparameters as we utilised the default settings provided by \citet{peft}.
Unless otherwise specified, a total of 30 demonstrations were used along with Static partitioning. 
%Both \emph{scalar} and \hypernet were tuned for 5 epochs. 
Both baselines (when applicable) and \moicl are tuned for 5 epochs. 
We use greedy decoding for generating from \moicl. The hyper-networks used in our experiments have parameters of 16M (t5-efficient-tiny), 31M (t5-efficient-mini), 60M (t5-small), and 220M (t5-base), respectively.
\subsection{Implementation Details}
\label{appendix:imple_details}
\paragraph{\moicl}
For all datasets used in the experiments, we fine-tuned all the \moicl weights and hyper-networks on the training set and evaluated them on the validation/development set at each epoch, selecting the ones with the highest performance. The results reported in all experiments were measured on the test set. For \emph{scalar}, we first sampled $D$ from the training set based on the different seeds and used the remaining training instances as $D_{T}$ (\S \ref{sec:moicl}). In experiments across multiple models (\S \ref{ssec:moicl_proxy}), the same fixed in-context demonstration set ($n=6, k=1$) was provided to all three models (Llama-3.2-1B, Llama-3.2-1B-Instruct, and Llama-3.2-3B), forming three separate experts. The outputs of these experts were then combined using a learned weighting function to generate the final output. For \hypernet in \S \ref{ssec:hypernet}, $D$ is not available during training and is used only during evaluation. We further separate $D_{T}$ into $D_{pool}$ and $D_{pair}$ randomly at each epoch, where demonstrations are sampled from $D_{pool}$ and $(x, y) \in D_{pair}$. While any model that produces weights can be used for the hyper-network, we attach a linear layer on top of a pre-trained encoder-decoder T5-small~\citep{2020t5} model. For the data efficiency analysis on \cref{fig:figure_data_eff} in \S \ref{ssec:data_eff}, we applied the same training step (10,240) to all different \moicl settings. 

\paragraph{Baselines}
For PEFT fine-tuned on RTE, we applied early stopping based on the dev set accuracy, as we observed that the training process was highly unstable. Both Ensemble-based ICL~\citep{min-etal-2022-noisy} and LENS~\cite{li2023finding} used the Direct method instead of the Channel method, which also applied for \moicl as well. For LENS, We first apply \emph{Progressive Example Filtering} to select 30 demonstrations, then perform \emph{Diversity-Guided Search} to obtain five permutations of the examples and report the average and standard deviation based on these five permutations. For \citet{le-etal-2022-shot}, we used SBERT model with \textit{all-roberta-large-v1} checkpoint\footnote{\url{https://www.sbert.net/docs/sentence_transformer/pretrained_models.html}} to calculate input–demonstration similarity, following the original setting. For LARA~\cite{huang2025divide}, we used the original implementation provided by the authors\footnote{\url{https://github.com/Chengsong-Huang/LARA}}, set $n$=$k$=30 (1-shot), increased the number of iterations from 20 to 40 (considering that switching from 2-shots to 1-shot increases the number of weights to learn), and modified the training instances to be a separate $n$×3 rather than $n$.

\section{Additional Analyses}

\subsection{Impact of Model Size}
\label{sec:model_size}

\begin{table}[t]
\resizebox{\linewidth}{!}{
\begin{tabular}{@{}lccc}
\toprule
\textbf{Method $\downarrow$ Model $\rightarrow$}   & \textbf{ll2-chat-7b} & \textbf{ll2-chat-13b} & \textbf{ll2-chat-70b} \\
\midrule
Concat-based ICL  & 73.09\std{3.21}     & 63.09\std{3.85}         &  69.42\std{1.78}       \\
\midrule
\moicl             &                     &                      &                      \\
\quad\emph{uniform} &  \textbf{79.35\std{0.22}}       &  63.60\std{1.84}       &  67.88\std{1.03}       \\
\quad\emph{scalar}  &  79.16\std{0.60}       &  \textbf{80.49\std{1.01}}       &  \textbf{82.26\std{0.65}}     \\
\bottomrule
\end{tabular}
}
\caption{Comparison on the TweetEval Offensive Test set across different sizes of the Llama-2 models. 
}
\label{table:model_size}
\end{table}

Considering the ongoing trend of scaling up LLMs, it is essential to analyse how the proposed method is affected by model size. In ~\cref{table:model_size}, we compare the accuracy of our proposed method on the TweetEval Offensive task when using Llama-2-chat models in various sizes (7B, 13B, 70B) as the target LLM. \footnote{We chose LLaMA-2-chat because it offers three clearly separated model sizes, making it suitable for analysing trends across scale. In contrast, LLaMA-3 currently provides only 8B and 70B models (while LLaMA-3.1 includes a 405B version, it was beyond our available computational resources)} Although the performance of the Llama-2-7B-chat model is somewhat unusual compared to the other two models, we observed that \moicl consistently outperforms concat-based ICL across all three model sizes.

\begin{table}[t]
\small
\resizebox{\linewidth}{!}{
\begin{tabular}{@{}lcc}
\toprule
\textbf{Hyper-network Model}          & \textbf{Offensive}    & \textbf{Hate} \\
\midrule
t5-efficient-tiny (16M) & \text{69.32\std{2.07}} | \text{74.60\std{2.03}} & \text{67.32\std{0.66}} | \text{60.48\std{4.56}} \\
t5-efficient-mini (31M) & \text{68.50\std{2.01}} | \text{73.74\std{1.43}} & \text{66.00\std{1.51}} | \text{56.61\std{0.90}} \\
t5-small (60M)          & \textbf{71.01\std{1.09}} | \textbf{76.65\std{1.31}} & \textbf{70.20\std{1.53}} | \textbf{65.07\std{5.22}} \\
t5-base (220M)          & \text{69.14\std{1.01}} | \text{74.40\std{2.39}}  & \text{68.24\std{0.75}} | \text{63.23\std{4.51}} \\
\bottomrule
\end{tabular}
}
\caption{Comparison on the TweetEval Offensive/hate Dev|Test set using Llama-3-8b-Instruct as a target LLM across different sizes of the hyper-network. The numbers in parentheses indicate the number of parameters.}
\label{table:model_size_hyper}
\end{table}

We also analysed the impact of hyper-network model size on the performance of \hypernet weights. ~\cref{table:model_size_hyper} compares the dev/test set accuracy on the TweetEval hate/offensive task based on the size of the T5 model used as the hyper-network. We chose T5 because it is available in a wide range of model sizes, allowing us to systematically examine the tradeoff between model capacity and performance. Importantly, the hyper-network must remain significantly smaller than the underlying LLM to maintain overall efficiency. From analysing the dev set results, we found that even with a very small model size (16M–60M), the hyper-network performed reasonably well. Based on this analysis, we selected T5-small (60M) as the default hyper-network, as it offers a favourable balance between computational efficiency and performance, representing less than 1\% of the total size of the main model (8B).

\subsection{Partitioning a Demonstration set $D$.}
\label{ssec:partition}
In this work, we analyse the following partitioning strategies: \partstatic, \partrandom, and \textbf{BM25}.
\partstatic means partitioning $n$ demonstrations into $k$ subsets, with each subset containing $n/k$ demonstrations. The demonstrations are selected randomly, but the subset sizes are kept fixed (hence, “static”).
\partrandom refers to partitioning into $k$ subsets, but both the subset sizes and the demonstration assignments are randomised.
\textbf{BM25} apply $k$-NN clustering based on BM25 scores over the demonstrations \citep{robertson2009probabilistic}, producing $k$ semantically coherent subsets with variable sizes.

\begin{table}[t]
%\vspace*{-1em}
\resizebox{\linewidth}{!}{
\begin{tabular}{@{}lccc}
\toprule
\textbf{\moicl Method} & \partstatic & \partrandom & \textbf{BM25} \\
\midrule
\emph{uniform}       &        &         &               \\
\quad $k=3$      &  \textbf{74.86\std{1.84}}          & \textbf{74.74\std{1.90}}  & \textbf{74.79\std{1.79}}   \\
\quad $k=5$      &  73.77\std{1.60}          & 74.09\std{1.35}     & 73.47\std{2.19}   \\
\quad $k=10$     &  74.00\std{0.87}           & 73.37\std{0.94}     & 74.40\std{0.82}   \\
\midrule
\emph{scalar}        &       &          &               \\
\quad $k=3$      &  76.14\std{1.48}         & 77.37\std{1.97}     & 77.21\std{2.02}   \\
\quad $k=5$      &  78.35\std{1.49}           & 77.67\std{2.69}     & 78.37\std{1.62}   \\
\quad $k=10$     &   \textbf{79.42\std{1.48}}          & \textbf{78.72\std{0.87}}     & \textbf{79.70\std{1.32}}  \\
\bottomrule
\end{tabular}
}
\caption{Analysis of partitioning methods on TweetEval Offensive dataset. Random and BM25 represent random clustering and clustering based on BM25 scores, respectively. 
Bold text signifies the highest accuracy for each method. 
}
\label{table:partitioning}
\end{table}

Table \ref{table:partitioning} compares the performance of \moicl methods and different partitioning methods (Static, Random, BM25) for the same $k$ (number of subsets). In \emph{uniform}, there is little difference between Static and Random and only a slight performance improvement with BM25. However, there is a common performance enhancement when \moicl \emph{scalar} are applied. 
This indicates that our proposed method is not significantly affected by partitioning methods and can be applied in a complementary manner across them. 
As such, we decided to use only the Static method in the other experiments. 

\begin{table}[t]
\resizebox{\linewidth}{!}{
\begin{tabular}{lccc}
\toprule
\textbf{\moicl} $\downarrow$ \textbf{Label Ratio} $\rightarrow$     & \textbf{27:3} & \textbf{25:5} & \textbf{23:7} \\
\midrule
\emph{scalar} ($n$=30, $k$=10)  &               &               &               \\
- Static Partitioning & 73.33\std{0.96}         & 72.72\std{0.76}         & 73.81\std{2.25}         \\
- BM25 Partitioning   & 74.63\std{1.03}         & 74.09\std{1.07}         & 74.42\std{1.50}  \\
\bottomrule
\end{tabular}
}
\caption{Performance comparison of static and BM25-based partitioning under varying levels of label imbalance on the TweetEval Offensive dataset with $k=10$ subsets. Each imbalance setting denotes the number of demonstrations per label (e.g., 27:3 indicates 27 for the majority label and 3 for the minority).
}
\label{table:imbal_partitioning}
\end{table}

To further assess the robustness of partitioning strategies, we investigate their behaviour under label imbalance (\S \ref{sec:Imbalance}). Specifically, we compare static and BM25-based partitioning on the TweetEval Offensive dataset using $k=10$ subsets, with varying degrees of label imbalance across demonstrations (from 27:3 to 23:7 ratios). As shown in Table~\ref{table:imbal_partitioning}, BM25 consistently outperforms the static strategy, especially in highly imbalanced settings such as 27:3 and 25:5. We attribute this to BM25’s ability to group demonstrations based on semantic similarity, which often correlates with label similarity. This clustering effect enables \moicl to assign more targeted and informative weights to subsets, thereby mitigating the negative impact of label imbalance. These findings suggest that while static partitioning is effective and stable in general, semantically informed strategies like BM25 can offer clear advantages when label distributions are skewed.

\subsection{Logits vs. Probabilities for Mixing Experts}
\label{appendix:mixing}

As stated in \S \ref{sec:moicl}, we mix the experts in the log domain. However, it is also possible—and perhaps more appropriate—to use a regular mixture of probabilities, as in \cref{eq:moicl_prob}. 
\begin{equation} \label{eq:moicl_prob}
p\left(y \mid D, x\right) \propto \left[\sum_{i=1}^{k} w_i p\left(y\mid D_i,x\right)\right]
\end{equation}
Accordingly, in \cref{fig:figure3}, we compare the accuracy trends based on partitioning size when using weighting in the probability and logit domains. In \emph{uniform}, whether logits or probabilities were used did not make a significant difference, but in \emph{scalar}, the impact was substantial. This is likely because distinct differences in the distribution patterns among experts (and thus useful information in the mixture) get diluted during the normalisation process when using probabilities.

\subsection{\moicl in a Generation Task}
\label{appendix:genration}

% \begin{table}[ht]
% \resizebox{\linewidth}{!}{
% \begin{tabular}{@{}lcc}
% \toprule
% \textbf{Methods ($n=12$)}            & \textbf{NQ-open (EM)} & \textbf{GSM-8K (Acc.)}\\
% \midrule
% Concat-based ICL   & \textbf{40.34\std{0.26}} &  61.83\std{1.77}        \\
% %Ensemble-based ICL \citep{min-etal-2022-noisy} & 0.3753\\
% %Random Search      & 0.4008  \\
% %\midrule
% Mixture of ICL (\emph{uniform})  & \\
% \quad $k = 6$            & 38.57\std{0.38} &  62.32\std{1.25}\\
% \quad $k = 12$           & 37.21\std{0.27} &  61.74\std{0.78}\\
% Mixture of ICL (\emph{scalar})  &\\
% \quad $k = 6$            & 38.83\std{0.38} &  63.20\std{0.50}\\
% \quad $k = 12$              & 38.33\std{0.25} &  62.14\std{1.03}\\
% Mixture of ICL (\emph{Hyper-network}) &\\
% \quad $k = 6$            & 38.90\std{0.19} &  \textbf{63.96\std{2.00}}\\
% \quad $k = 12$              &  &  61.49\std{2.19}\\
% \bottomrule
% \end{tabular}
% }
% \caption{Comparison between baseline methods and \moicl on NQ-open and GSM-8K using Llama-3-8B. $k$ represents the number of demonstrations subset, where the total number of demonstrations ($n$) is 12}
% \label{table:generation}
% \end{table}

\begin{table}[t]
%\resizebox{\linewidth}{!}
%{
\begin{tabular}{lc}
\toprule
\textbf{Methods ($n=12$)}            & \textbf{GSM8K (Acc.)}\\
\midrule
Concat-based ICL   &   61.83\std{1.77}        \\
%Ensemble-based ICL \citep{min-etal-2022-noisy} & 0.3753\\
%Random Search      & 0.4008  \\
%\midrule
Mixture of ICL (\emph{uniform})  & \\
\quad $k = 6$            &   62.32\std{1.25}\\
\quad $k = 12$           &   61.74\std{0.78}\\
Mixture of ICL (\emph{scalar})  &\\
\quad $k = 6$            &   63.20\std{0.50}\\
\quad $k = 12$              &   62.14\std{1.03}\\
Mixture of ICL (\hypernet) &\\
\quad $k = 6$            &   \textbf{63.96\std{2.00}}\\
\quad $k = 12$           &  61.49\std{2.19}\\
\bottomrule
\end{tabular}
%}
\caption{Comparison between baseline methods and \moicl on GSM8K using Llama-3-8B. $k$ represents the number of demonstrations subset, where the total number of demonstrations ($n$) is 12}
\label{table:generation_gsm}
\end{table}

\begin{table}[t]
\resizebox{\linewidth}{!}{
\begin{tabular}{lc}
\toprule
\textbf{Methods}            & \textbf{NQ-open (EM)}\\
\midrule
Concat-based ICL & \\
\quad $n=12$   &   40.34\std{0.26}      \\
\quad $n=24$   &   40.58\std{0.47}      \\
\quad $n=48$   &   40.07\std{0.50}      \\
\quad $n=96$   &   -      \\
\midrule
Concat-based ICL (Llama-3.1-8B) & \\
\quad $n=12$   &   39.70\std{0.56}      \\
\quad $n=24$   &   40.29\std{0.75}      \\
\quad $n=48$   &   40.17\std{0.58}      \\
\quad $n=96$   &   40.47\std{0.68}      \\
\midrule
%Ensemble-based ICL \citep{min-etal-2022-noisy} & 0.3753\\
%Random Search      & 0.4008  \\
%\midrule
% Mixture of ICL (\emph{uniform}, $n=96$)  & \\
% \quad $k = 6$            &   40.86\std{0.31} \\
% \quad $k = 12$           &   40.67\std{0.20} \\
Mixture of ICL (\emph{scalar}, $n=96$)  &\\
\quad $k = 6$            &   \textbf{40.86\std{0.31}} \\
\quad $k = 12$              &   40.74\std{0.35}\\
Mixture of ICL (\hypernet, $n=96$) &\\
\quad $k = 6$            &  40.66\std{0.26} \\
\quad $k = 12$           &  40.64\std{0.48}\\
\bottomrule
\end{tabular}
}
\caption{Comparison between baseline methods and \moicl on NQ-open using Llama-3-8B. We also use Llama-3.1-8B for concat-based ICL to evaluate the effectiveness of a longer-context model. $k$ represents the number of demonstrations subset, where the total number of demonstrations is $n$. For $n = 96$, the performance of concat-based ICL with Llama-3-8B could not be measured due to exceeding the context length limit.}
\label{table:generation_NQ}
\end{table}

In addition to the classification tasks in ~\S \ref{ssec:classification_tasks}, we also apply our \moicl on a generation task, NQ-open~\citep{lee-etal-2019-latent} and GSM8K~\citep{cobbe2021gsm8k}, in ~\cref{table:generation_gsm} and ~\cref{table:generation_NQ}. However, unlike in classification tasks, \moicl did not show significant EM improvements over baseline approaches in NQ. 
For GSM8K, although \moicl using \emph{scalar} weights and \hypernet weights achieves performance improvements over Concat-based ICL and uniform weighting, these gains are not statistically significant when considering the standard deviations.
Nevertheless, as seen in ~\S \ref{sec:noised_demon}, \moicl exhibited strong robustness in situations involving perturbed demonstrations, proving the usefulness of the expert's tuned weights.

\subsection{Reducing Training Costs Through Caching}
\label{appendix:training_cache}

While \moicl does not fine-tune the expert models themselves, it leverages their next-token distributions to guide the weighting function. A key efficiency gain arises from caching these distributions during the training process. Because the underlying LLM parameters remain fixed, the output distributions computed at the outset can be stored and subsequently reused for all training instances. This approach eliminates the need to recompute the distributions from scratch for each training epoch.

\begin{table*}[ht]
\resizebox{\linewidth}{!}{
\begin{tabular}{@{}lcc}
\toprule
& \bf{Epoch w/o cache (1st epoch)} & \bf{Epoch w/ cache (2nd-last epoch)}    \\
\midrule
Training FLOPs per Instance           & 21.61 TFLOPs                  & 205 FLOPs   \\
Training FLOPs per Epoch              & 216.15 TFLOPs                 & 2,050 FLOPs \\
\midrule
Inference FLOPs per Instance          & 22.07 TFLOPs                  & -                 \\
Inference FLOPs in Total              & 18,975.90 TFLOPs              & -                 \\
\midrule
FLOPs in total (Training + Inference) & 19,191.91 TFLOPs              & -    \\
\bottomrule
\end{tabular}
}
\caption{Training and inference FLOPs for \moicl on the offensive classification task. We report per-instance and per-epoch costs both with and without caching, using a \emph{scalar} weighting function, $k$=10, and 10 demonstrations for weight tuning.}
\label{table:cache}
\end{table*}

~\cref{table:cache} reports the number of floating-point operations (FLOPs) for \moicl’s training and inference on the offensive classification task, using a \emph{scalar} weighting function, $k$=10, and 10 demonstrations for weight tuning\footnote{In this setup, we use $k$=10 and 10 demonstrations (20 annotated demonstrations in total) for weight tuning, as these values have been found (in \S \ref{ssec:data_eff} and \cref{fig:figure_data_eff}) to provide a practical balance between performance and computational cost.}. Without caching, the first training epoch requires 21.61 TFLOPs per instance and 216.15 TFLOPs per epoch. By contrast, once caching is enabled, training costs drop drastically to approximately 205 FLOPs per instance and 2,050 FLOPs per epoch. Moreover, these training costs represent only about 1\% of the total FLOPs when inference—requiring roughly 18,975.90 TFLOPs—is considered. By quantifying training costs in terms of FLOPs, we clearly demonstrate that freezing the LLM and leveraging caching yields substantial efficiency gains, making \moicl highly practical.

\subsection{Applying \moicl to Black-Box Models}
\label{appendix:black_box}

Proprietary LLMs offered via API often differ in the level of access they provide to their underlying distributions. While some grant full distributional access, others restrict output to the top-k logits, enable user-defined logit biases, or limit the user to text-only outputs without any direct probability information. Such constraints pose challenges for methods like \moicl, which rely on next-token distributions to tune the weighting function.

However, recent work has shown that extracting logits or other internal model components from proprietary, restricted-access models is feasible under a range of conditions. \citet{morris2024language} detail approaches to language model inversion, while \citet{carlinistealing} and \citet{finlayson2024logits} explore methods for partially revealing models’ logits and other parameters despite limited access. These techniques demonstrate that even under scenarios with partial distributional access, user-specified logit biases, or text-only responses, there are practical strategies to approximate or recover critical information about the model’s outputs.

To investigate the applicability of \moicl in settings where only top-k logits are available, in \cref{table:black-box}, we trained and evaluated \moicl on the TweetEval Offensive and GSM8K tasks under a setting where only the top 40 logits were provided. \moicl demonstrates its performance under a black-box setting—where only partial distributional information is accessible—demonstrating its potential adaptability to restricted-access environments.

\begin{table}[t]
\resizebox{\linewidth}{!}{
\begin{tabular}{@{}lcc}
\toprule
                     & \bf{Offensive}    & \bf{GSM8K}       \\
\midrule
\moicl (\emph{scalar}, $n=k$)  & 81.33\std{0.69} & 62.14\std{1.03} \\
w/ black-box setting & 79.81\std{0.39} & 62.26\std{1.12} \\
\bottomrule
\end{tabular}
}
\caption{\moicl on the TweetEval Offensive and GSM8K tasks using Llama-3-8B-Instruct under a black-box scenario where only the top-40 logits are accessible.}
\label{table:black-box}
\end{table}

\subsection{\moicl with Preference Fine-Tuning}
\label{appendix:DPO}

\begin{table}[t]
\resizebox{\linewidth}{!}{
\begin{tabular}{@{}lc}
\toprule
& \bf{HH-RLHF Harmless}     \\
%& (Acc.) \\
\midrule
Concat-based ICL (n=8) & 55.02 \\
\midrule
\moicl ($n=2, k=4$) & \\
%\midrule
\quad \emph{uniform}  & 55.41  \\
\quad \hypernet (chosen only)  & 56.23  \\
\quad \hypernet (chosen + rejected)  & \textbf{57.14}  \\
\bottomrule
\end{tabular}
}
\caption{Results (accuracy) of \moicl using Direct Preference Optimization (DPO) on the HH-RLHF Harmless dataset test set using Llama-3-8B-Instruct. "chosen + rejected" represents chosen and rejected demonstrations assigned to experts ($k=2$) and anti-experts ($k=2$), respectively.}
\label{table:DPO}
\end{table}

One notable feature of \moicl is its ability to incorporate anti-experts into ICL, as discussed in \S \ref{sec:anti_expert}. This opens up the possibility of applying \moicl to preference fine-tuning (\eg for RLHF) where, for a given input query, there are preferred responses (“chosen”) and dispreferred responses (“rejected”). \moicl can learn the weighting function of experts and anti-experts through Direct Preference Optimization~\cite[DPO;][]{rafailov2024direct} by assigning "chosen" demonstrations to experts and "rejected" demonstrations to anti-experts. 

\cref{table:DPO} compares the results of training and evaluating \moicl DPO on the HH-RLHF harmless dataset \cite{bai2022training} with concat-based ICL and \emph{uniform} weighting. When “chosen” and “rejected” demonstrations were assigned to experts and anti-experts, respectively (chosen + rejected), \moicl outperforms concat-based ICL by 2.1\%. The learned weights of \moicl showed a clear distinction, with experts having weights of 0.43 and 0.60, while anti-experts had weights of 0.15 and -0.32. This result demonstrates that \moicl can successfully incorporate anti-experts into ICL, proving its potential for extension to preference tuning.

\subsection{Comparison with Longer Context Models}
\label{appendix:longer_context}
By distributing demonstrations across experts or using them for tuning the weight function, \moicl can mitigate the limitations imposed by the model’s context length. However, this limitation could also be addressed by using an LLM with a longer context length. To evaluate whether \moicl remains effective when compared to longer-context LLMs, we assess the performance of longer-context LLMs (Llama-3.1-8B and Llama-3.1-8B-Instruct with a 128k context length) to \moicl based on the original base LLMs (Llama-3-8B and Llama-3-8B-Instruct with an 8k context length) in scenarios where the context length of the base LLMs was exceeded. 

\begin{figure}[t]
    \centering

    \begin{subcaptionbox}{TweetEval Offensive\label{fig:figure_data_eff_off_ll31}}{
        \includegraphics[width=\linewidth]{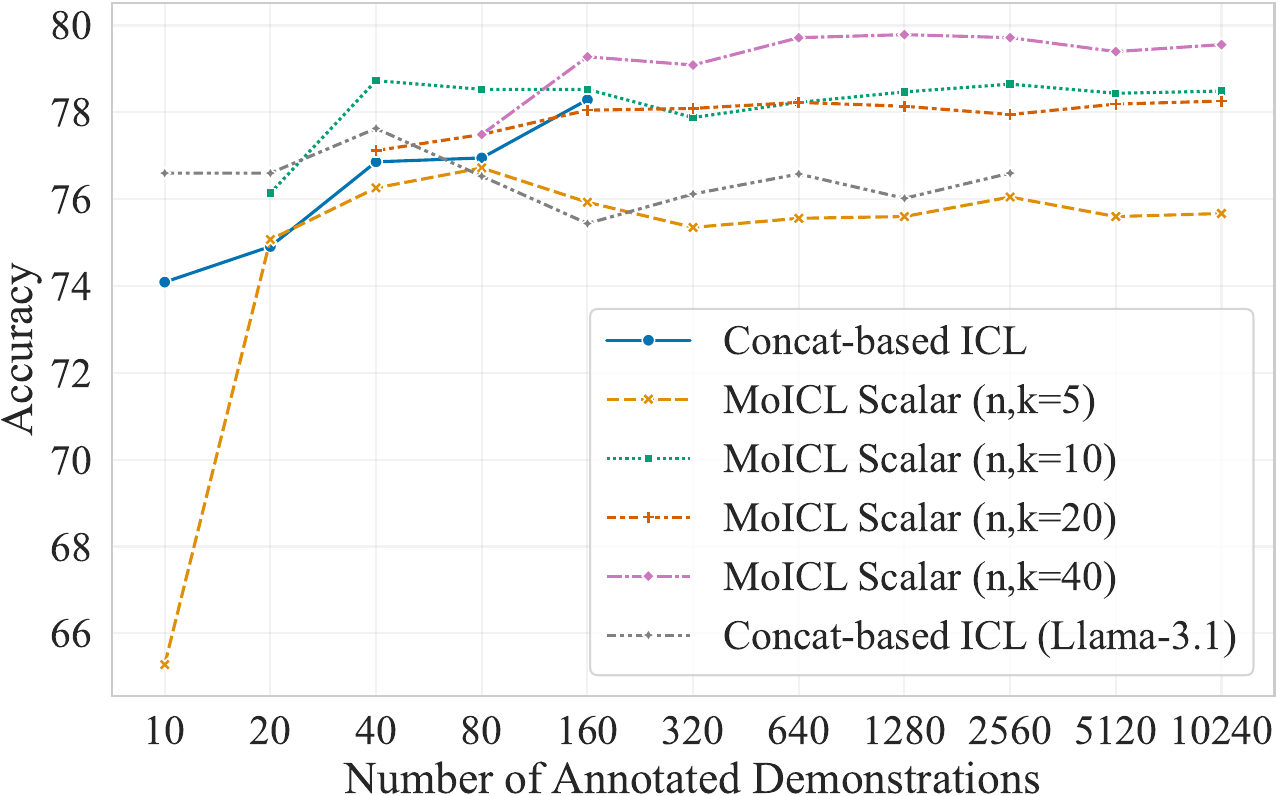}
    }
    \end{subcaptionbox}

    \begin{subcaptionbox}{TweetEval Hate\label{fig:figure_data_eff_hate_ll31}}{
        \includegraphics[width=\linewidth]{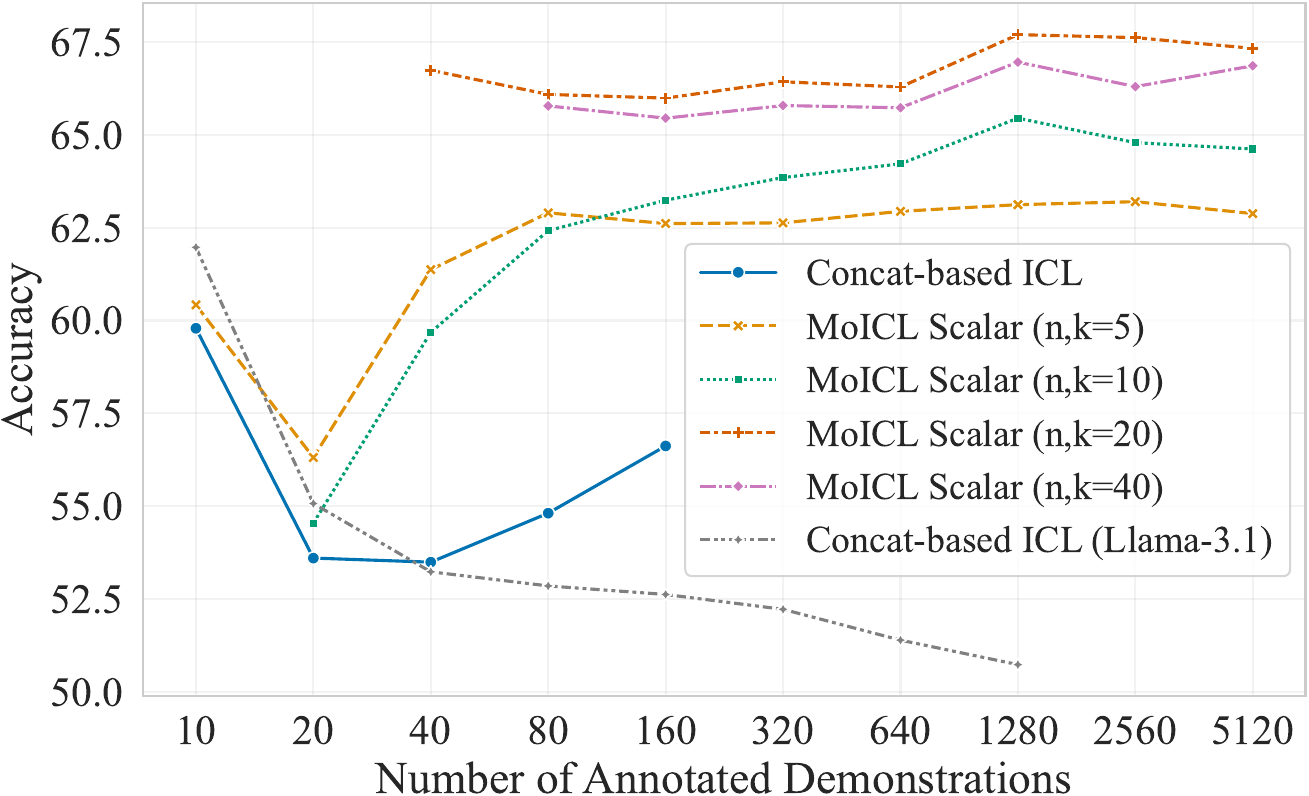}
    }
    \end{subcaptionbox}
    
    \caption{An analysis of \moicl's data-efficiency on the TweetEval offensive/hate test set using Llama-3-8B-Instruct. Concat-based ICL concatenated all available demonstrations ($x$-axis), exceeding the context length when ($n>160$). We also use Llama-3.1-8B-Instruct for concat-based ICL to evaluate the effectiveness of a longer-context model.} 
    %
    %\moicl \emph{scalar} ($k=n$) assigned the designated demonstrations to the experts while using the remaining available demonstrations for fine-tuning.}
    \label{fig:figure_data_eff_ll31}
\end{figure}

As shown in \cref{table:generation_NQ}, on NQ-open, Llama-3-8B model fails to produce an output when $n = 96$ due to context length limitations. However, \moicl efficiently distributes demonstrations across experts ($k = 6$, $k = 12$), achieving effective performance that even outperforms the longer-context model, Llama-3.1-8B. Similarly, as shown in \cref{fig:figure_data_eff_ll31} for the TweetEval Offensive and Hate datasets, Llama-3-8B-Instruct model fails to produce an output when $n > 160$ due to context length limitations. While the longer-context model, Llama-3.1-8B-Instruct, maintained reasonable performance even at higher $n$ values (except when results could not be obtained due to memory limitations at $n > 2,560$ for Offensive and $n > 1,280$ for Hate), \moicl outperforms the longer-context model on both datasets. 

One possible explanation for this is that simply increasing the number of demonstrations in ICL does not always lead to better performance, as performance may saturate or even degrade \cite{chen2023how}. In this regard, \moicl’s effectiveness compared to longer-context models may be attributed to its more efficient ICL framework, which distributes demonstrations across experts and utilises them during training.

\section{Prompt Templates}

~\cref{table:prompt} presents the corresponding metric and prompt template for all tasks included in the experiments. For NQ, the delimiter for ICL demonstrations was `\textbackslash{}n\textbackslash{}n'. For the remaining tasks, `\textbackslash{}n' was used as the delimiter.

\begin{table*}
%\vspace*{-1em}
\caption{Prompt template setting details for the tasks. The double curly braces "\{\{\}\}" signify input data.}
\label{table:prompt}
\small
\resizebox{\textwidth}{!}{
\begin{tabular}{lcl}
\textbf{Task} &
  \textbf{Metric} &
  \multicolumn{1}{l}{\textbf{Prompt Template}} \\
  \toprule
TweetEval Offensive &
  Accuracy &
  \begin{tabular}[c]{@{}l@{}}Classify tweets that are offensive as offensive, and tweets that are not offensive as neutral.\\ \{\{\textit{ICL Demonstrations}\}\}\\ Tweet: \{\{\textit{tweet}\}\}\\ Label: \end{tabular} \\
\midrule
TweetEval Hate &
  Accuracy &
  \begin{tabular}[c]{@{}l@{}}Classify tweets that are hateful against immigrants or women as hate and tweets that are \\not hateful against immigrants or women as neutral.\\ \{\{\textit{ICL Demonstrations}\}\}\\ Tweet: \{\{\textit{tweet}\}\}\\ Label: \end{tabular} \\
\midrule
SST2 &
  Accuracy &
  \begin{tabular}[c]{@{}l@{}}Classify sentences that are negative as negative and sentences that are positive as positive.\\ \{\{\textit{ICL Demonstrations}\}\}\\ Sentence: \{\{\textit{sentence}\}\}\\ Label: \end{tabular} \\
\midrule
RTE &
  Accuracy &
  \begin{tabular}[c]{@{}l@{}}Classify two sentences that entail each other as true and two sentences that do not \\entail each other as false.\\ \{\{\textit{ICL Demonstrations}\}\}\\ Sentence1: \{\{\textit{first sentence}\}\} Sentence2: \{\{\textit{second sentence}\}\}\\ Label: \end{tabular} \\
\midrule
FEVER &
  Accuracy &
  \begin{tabular}[c]{@{}l@{}}Classify claims that are false as refuted, and tweets that are true as supported.\\ \{\{\textit{ICL Demonstrations}\}\}\\ Claim: \{\{\textit{claim}\}\}\\ Label: \end{tabular} \\
\midrule
PAWS &
  Accuracy &
  \begin{tabular}[c]{@{}l@{}}Classify the two sentences as yes if they are paraphrases of each other, and if not, \\classify them as no.\\ \{\{\textit{ICL Demonstrations}\}\}\\ sentence1: \{\{\textit{first sentence}\}\} sentence2: \{\{\textit{second sentence}\}\}\\ label: \end{tabular} \\
\midrule
QNLI &
  Accuracy &
  \begin{tabular}[c]{@{}l@{}}Classify as yes if the sentence contains the answer to the question, if not, classify as no.\\ \{\{\textit{ICL Demonstrations}\}\}\\ sentence: \{\{\textit{sentence}\}\}\\ question: \{\{\textit{question}\}\}\\ label: \end{tabular} \\
\midrule
NQ &
  EM &
  \begin{tabular}[c]{@{}l@{}}\{\{ICL Demonstrations\}\}\\ title: \{\{\textit{title}\}\} text: \{\{\textit{text}\}\}\\ Question: \{\{\textit{question}\}\}\\ Answer: \end{tabular} \\
\midrule
GSM8K &
  EM &
  \begin{tabular}[c]{@{}l@{}}\{\{ICL Demonstrations\}\}\\ Q: \{\{\textit{question}\}\}\\ A: \{\{CoT \textit{reasoning}\}\} The answer is \end{tabular} \\
\midrule
HH-RLHF &
  Accuracy &
  \begin{tabular}[c]{@{}l@{}}\{\{ICL Demonstrations\}\}\\ \{\{\textit{Chosen response}\}\}\\ or \\\{\{\textit{Rejected response}\}\}  \end{tabular} \\
\bottomrule
\end{tabular}
}
\end{table*}

\section{Computation Details}
The experiments were conducted using NVIDIA A100 40GBs and 80GBs with 120GB of RAM. The GPU hours vary depending on the models and tasks; tuning \moicl scalar weights ($n,k=30$) on TweetEval offensive takes approximately 1 hour and 20 minutes per epoch.

\end{document}